\documentclass[runningheads]{llncs}
\usepackage[T1]{fontenc}
\usepackage{graphicx}
\usepackage{booktabs}
\usepackage[misc]{ifsym}
\newcommand{\corr}{(\Letter)}

\usepackage{amsthm}
\usepackage{amsmath}

\usepackage[utf8]{inputenc} 
\usepackage[T1]{fontenc}    
\usepackage{hyperref}       
\usepackage{url}            
\usepackage{booktabs}       
\usepackage{amsfonts}       
\usepackage{nicefrac}       
\usepackage{microtype}      
\usepackage{xcolor}         
\usepackage{amsmath}
\usepackage{amssymb}
\usepackage[ruled]{algorithm2e}

\usepackage{subcaption}
\usepackage{dsfont}
\usepackage{fontawesome5}
\usepackage{bbding}
\usepackage[disable]{todonotes}

\usepackage{pgfplots}
\pgfplotsset{compat=1.17}
\usetikzlibrary{intersections} 
\usepgfplotslibrary{fillbetween} 

\usepackage{tikz}
\usetikzlibrary{shapes.geometric, arrows}
\usetikzlibrary{arrows.meta, positioning, quotes}
\usetikzlibrary{calc}
\usetikzlibrary{automata}
\usetikzlibrary{tikzmark}

\newcommand{\mathset}[1]{\{ #1 \}}

\newcommand{\indicatorNonset}[1]{\ensuremath{\mathds{1}_{#1}}}
\newcommand{\reals}{\mathbb{R}}
\newcommand{\naturals}{\mathbb{N}}

\newcommand{\mdp}[1]{\ensuremath{\mathcal{#1}}}
\newcommand{\mdpGamma}{\ensuremath{\gamma}}
\newcommand{\mdpDiscount}{\ensuremath{\gamma}}
\newcommand{\mdpLabel}{\ensuremath{L}}
\newcommand{\mdpCommonLabel}{\ensuremath{\ell}}
\newcommand{\mdpStates}{\ensuremath{S}}
\newcommand{\mdpCommonState}{\ensuremath{s}}
\newcommand{\mdpInit}{\ensuremath{\mdpCommonState_{\init}}}
\newcommand{\mdpActions}{\ensuremath{A}}
\newcommand{\mdpCommonAction}{\ensuremath{a}}
\newcommand{\mdpTrans}{\ensuremath{p}}
\newcommand{\mdpReward}{\ensuremath{R}}
\newcommand{\mdpCommonReward}{\ensuremath{r}}

\newcommand{\policy}{\ensuremath{\pi}}

\newcommand{\valueFunction}{\ensuremath{V}}

\newcommand{\init}{\ensuremath{I}}
\newcommand{\dfa}[1]{\ensuremath{\mathsf{#1}}}

\newcommand{\dfaStates}{\ensuremath{Q}}
\newcommand{\dfaStatesOf}[1]{\ensuremath{\dfaStates^{\dfa{#1}}}}
\newcommand{\dfaCommonState}{\ensuremath{q}}

\newcommand{\dfaInitState}{\ensuremath{\dfaCommonState_{\init}}}
\newcommand{\dfaTrans}{\ensuremath{\delta}}
\newcommand{\dfaAcc}{\ensuremath{F}}
\newcommand{\dfaAccOf}[1]{\ensuremath{\dfaAcc^{\dfa{#1}}}}
\newcommand{\languageOf}{\ensuremath{\mathcal{L}}}

\newcommand{\propInput}{\ensuremath{\ell}}

\newcommand{\rewardMachine}[1]{\ensuremath{\mathcal{#1}}}
\newcommand{\rmStates}{\ensuremath{U}}
\newcommand{\rmStatesOf}[1]{\ensuremath{\rmStates^\rewardMachine{#1}}}
\newcommand{\rmCommonState}{\ensuremath{u}}
\newcommand{\rmCommonStateOf}[1]{\ensuremath{u^{\rewardMachine{#1}}}}
\newcommand{\rmInit}{\ensuremath{\rmCommonState_{\init}}}
\newcommand{\rmInitOf}[1]{\ensuremath{\rmInit^{\rewardMachine{#1}}}}
\newcommand{\rmAcc}{\ensuremath{\dfaAcc}}
\newcommand{\rmAccOf}[1]{\ensuremath{\dfaAcc^{\rewardMachine{#1}}}}
\newcommand{\rmTrans}{\ensuremath{\delta}}

\newcommand{\rmOut}{\ensuremath{\sigma}}

\newcommand{\gridGoal}{\faSave{}}

\newcommand{\textPipe}{\ensuremath{P}}
\newcommand{\textDoor}{\ensuremath{D}}
\newcommand{\textGenerator}{\ensuremath{G}}

\newcommand{\iconPipe}{\faShower{}}
\newcommand{\iconDoor}{\faUnlock{}}
\newcommand{\iconGenerator}{\faPlug{}}

\newcommand{\ltlf}{\ensuremath{\text{LTL}_f}}

\newcommand{\causalDiagram}{\ensuremath{\mathcal{C}}}

\newcommand{\nextOp}{\ensuremath{\textbf{X}}}
\newcommand{\untilOp}{\ensuremath{\textbf{U}}}
\newcommand{\weakUntilOp}{\ensuremath{\textbf{W}}}

\newcommand{\globallyOp}{\ensuremath{\textbf{G}}}

\newcommand{\eventFormula}{\ensuremath{\psi_{\marlEventSet}}}

\newcommand{\marlEventSet}{\ensuremath{\Sigma}}
\newcommand{\marlEventAlphabet}{\ensuremath{2^{\marlEventSet}}}
\newcommand{\marlAtomicEvent}{\ensuremath{e}}

\newcommand{\marlProjection}{\ensuremath{P}}

\newcommand{\marlShared}{\ensuremath{I}}

\newcommand{\marlIndistinguishability}{\ensuremath{\sim}}

\newcommand{\marlEventSeq}{\xi}
\newcommand{\marlEventSeqs}[1]{\ensuremath{\marlAtomicEvent_1 \ldots e_{#1}}}

\newcommand{\parallelComposition}{\ensuremath{\|}}

\newcommand{\bisim}{\ensuremath{\cong}}

\newcommand{\bisimAlt}{\ensuremath{\rho}}

\newcommand{\rmCompDfa}[2]{\ensuremath{\rewardMachine{#1} \parallelComposition \dfa{#2}}}

\newcommand{\rmMdpTrajectory}[1]{%
  \ensuremath{%
    \mdpCommonState_0 \rmCommonState_0 \mdpCommonAction_0 \cdots
    \allowbreak
    \mdpCommonAction_{#1-1} \mdpCommonState_{#1} \rmCommonState_{#1}}}

\newcommand{\teamTrajectory}{\ensuremath{\mdpCommonState_0 \rmCommonState_0 \cdots \mdpCommonState_k \rmCommonState_k}}

\newcommand{\localStatesTrajectorySingle}{\ensuremath{\mdpCommonState_0^i \rmCommonState_0^i \cdots \mdpCommonState_k^i \rmCommonState_k^i}}

\begin{document}

\title{Decentralizing Multi-Agent Reinforcement Learning with Temporal Causal Information\thanks{Code available at \url{https://github.com/corazza/tcdmarl}}}

\titlerunning{Decentralized MARL with Temporal Causality}

\author{Jan Corazza\inst{1} \corr \and
      Hadi Partovi Aria\inst{2} \and
      Hyohun Kim\inst{2} \and
      Daniel Neider\inst{1} \and
      Zhe Xu\inst{2}}


\authorrunning{Jan Corazza et al.}

\institute{Research Center Trustworthy Data Science and Security of the University Alliance Ruhr, Department of Computer Science, TU Dortmund University, Germany \email{\{jan.corazza,daniel.neider\}@tu-dortmund.de} \and
      Arizona State University, USA \email{\{hpartovi, hkim450, xzhe1\}@asu.edu}}

\maketitle              

\begin{abstract}
      Reinforcement learning (RL) algorithms can find an optimal policy for a single agent to accomplish a particular task.
      However, many real-world problems require multiple agents to collaborate in order to achieve a common goal.
      For example, a robot executing a task in a warehouse may require the assistance of a drone to retrieve items from high shelves.
      In Decentralized Multi-Agent RL (DMARL), agents learn independently and then combine their policies at execution time, but often must satisfy constraints on compatibility of local policies to ensure that they can achieve the global task when combined.
      In this paper, we study how providing high-level symbolic knowledge to agents can help address unique challenges of this setting, such as privacy constraints, communication limitations, and performance concerns.
      In particular, we extend the formal tools used to check the compatibility of local policies with the team task, making decentralized training with theoretical guarantees usable in more scenarios.
      Furthermore, we empirically demonstrate that symbolic knowledge about the temporal evolution of events in the environment can significantly expedite the learning process in DMARL.

      \keywords{Temporal Causality  \and Multi-Agent Reinforcement Learning \and Reward Machines \and Formal Methods.}
\end{abstract}

\section{Introduction}
\label{sec:introduction}

One approach to solving a multi-agent problem is to learn a centralized policy that controls all agents simultaneously.
Such a centralized controller is conceptually straightforward, but realizing it is often impractical~\cite{kazemi2023assumeguarantee} if the number of agents is large or there is an imposed need for decentralization.
As the number of agents increases, the state space grows exponentially, making it less tractable to learn a centralized policy.
Decentralization is imposed when agents are physically separated, communication is limited, or privacy is a concern. Centralized policies, by contrast, often assume seamless communication, which is unrealistic in many real-world scenarios.
This challenge is further exacerbated by sparse reward signals with temporal dependencies, inherent in many real-world tasks.

Instead of learning a centralized policy, agents can learn decentralized policies that allow them to act independently and reduce the need for communication.
These independent policies are then combined at execution time to solve the team task.
To this end, Neary et al. \cite{neary_reward_2021} propose a DMARL algorithm called Decentralized Q-learning with Projected Reward Machines (DQPRM).
DQPRM decomposes a global team task specification into a set of local task specifications, one for each individual agent.
Agents are trained independently to learn policies based on their local task specifications.
To ensure that the agents' local policies, when combined, satisfy the overall team task, DQPRM enforces strict \textbf{compatibility criteria} between the local and team specifications.
In Section~\ref{sec:dmarl}, we provide a brief overview of DQPRM.

Reward machines (RMs, formalized in Definition~\ref{def:rm}) are deterministic, finite-state automata that transduce sequences of relevant high-level events (labels) into sequences of rewards, thereby capturing the temporal nature of a sparse reward signal, and serving as specifications for RL tasks.
Reward machines have been widely studied in literature \cite{DBLP:journals/corr/abs-2010-03950,Xu_2020,pmlr-v236-corazza24a,paliwal2023reinforcement,dohmen2022inferring,Azran}, but the primary focus has been on the single-agent case.
In multi-agent problems, communication limitations mean individual agents cannot sense all events, preventing them from accessing the full team state.
DQPRM frames this limitation in terms of projected (local) RMs, that capture a particular agent's contribution to the team goal (with respect to events that the agent can sense).

\begin{figure}[!htbp]
  \centering
  \begin{subfigure}{.48\columnwidth}
    \centering
    \scalebox{0.8}{\begin{tikzpicture}[scale=0.5]

    \draw[opacity=0.2] (0,0) grid (9,9); \draw[opacity=1] (0,0) rectangle (9,9);

    \fill[black!50] (3,3) rectangle (9,4);
    \fill[black!50] (0,5) rectangle (5,6);

    \fill[yellow!50] (4,6) rectangle (5,9);

    \foreach \y in {0,1,2} {
            \draw[->, blue, thick, line width=1pt] (3.5,\y+0.5) -- (2.5,\y+0.5);
        }

    \fill[green!10] (5,1) rectangle (6,2);
    \node at (5.5,1.5) {\faRobot};
    \node at (5.5,0.5) {(1)};

    \fill[green!10] (1,7) rectangle (2,8);
    \node at (1.5,7.5) {\faRobot};
    \node at (1.5,6.5) {(2)};

    \fill[red!10] (8,2) rectangle (9,3);
    \node at (8.5,2.5) {\large \iconPipe};
    \node at (8.5,1.5) {(\textPipe)};

    \fill[red!10] (7,7) rectangle (8,8);
    \node at (7.5,7.5) {\large \iconGenerator};
    \node at (7.5,6.5) {(\textGenerator)};

    \fill[yellow!50] (1,1) rectangle (2,2);
    \node at (1.5,1.5) {\iconDoor};
    \node at (1.5,0.5) {(\textDoor)};

\end{tikzpicture}}
    \caption{Environment}
    \label{fig:case-study-3-gridworld}
  \end{subfigure}%
  \hfill
  \begin{subfigure}{.48\columnwidth}
    \centering
    \scalebox{0.8}{\begin{tikzpicture}[>=stealth, node distance=1.5cm, every state/.append style={scale=0.7}, initial text=]
    \node[state, initial] (0) at (0, 0) {$\rmInit$};
    \node[state] (1) at (1, 1) {$\rmCommonState_1$};
    \node[state] (2) at (1, -1) {$\rmCommonState_2$};
    \node[state] (3) at (2, 0) {$\rmCommonState_3$};
    \node[state] (4) at (2, -2) {$\rmCommonState_4$};
    \node[state, accepting] (5) at (3.5, 0) {$\rmCommonState_5$};

    \path[->] (0) edge node[above left] {\textPipe} (1);
    \path[->] (0) edge node[below left] {\textDoor} (2);
    \path[->] (1) edge node[above right] {\textDoor} (3);
    \path[->] (2) edge node[below right] {\textPipe} (3);
    \path[->] (2) edge node[below left] {\textGenerator} (4);
    \path[->] (3) edge node[above] {\textGenerator} (5);
\end{tikzpicture}}
    \caption{Team Reward Machine}
    \label{fig:case-study-3-team-rm}
  \end{subfigure}
  \caption{\emph{Generator Task}}
  \label{fig:case-study-3}
\end{figure}
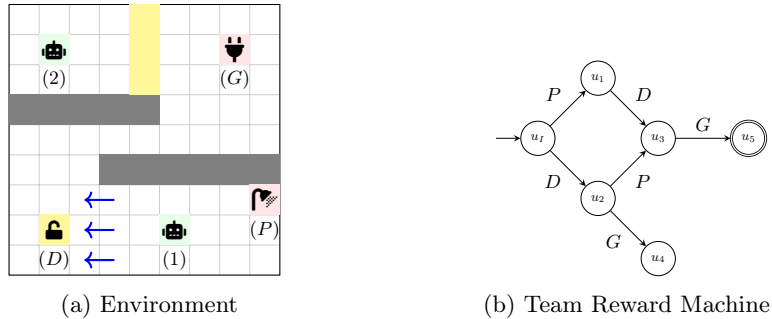

To illustrate the limitations of DQPRM's strict compatibility criteria, we will introduce a running example called the \emph{Generator Task}, depicted in Figure~\ref{fig:case-study-3}.
In this task, two agents move on a gridworld (Figure~\ref{fig:case-study-3-gridworld}), and must collaborate to power a generator during a flooding incident.

Agent $1$'s task is to prevent the flooding by closing the pipe (\iconPipe{}, \textPipe{}), and to unlock the door (\iconDoor{}, \textDoor{}) for agent $2$ (in any order). The task for agent $2$ is to wait for the door to be unlocked (\iconDoor{}), i.e., for the yellow barrier to disappear, and then power the generator (\iconGenerator{}, \textGenerator{}).
The team reward machine, depicted in Figure~\ref{fig:case-study-3-team-rm}, also specifies a failure condition: if agent $2$ is let into the room and powers the generator \emph{before} the flooding has been resolved, the team RM will enter an unsafe state, $\rmCommonState_4$, from which no further actions can result in a positive reward.
The agents obtain a reward of $1$ and end the task if their joint actions transition the team RM to into an accepting state, $\rmCommonState_5$, and $0$ in all steps prior.

The challenge in this example arises from the agents' limited communication about the task's status.
In particular, agent $1$ does not communicate whether it is safe to power up the generator (i.e., whether event \iconPipe{} has happened).
In other words, agent $2$ does not know whether the team is collectively in state $\rmCommonState_2$ (\iconGenerator{} is unsafe) or $\rmCommonState_3$ (\iconGenerator{} is safe and completes the task).
Although both agents have an individual strategy that solves the team task, this specification fails to meet the requirements of DQPRM, which strictly prohibit outcomes to depend on local event synchronization.

Recent work in causal reinforcement learning~\cite{paliwal2023reinforcement,pmlr-v236-corazza24a} proposes methods that leverage causal knowledge of the long-term temporal evolution of high-level environmental events to improve the exploration-exploitation trade-off.
Paliwal et al.~\cite{paliwal2023reinforcement} introduce Temporal Logic-based Causal Diagrams (TL-CDs), a formalism that captures such \emph{temporal-causal} knowledge by enabling experts to specify causal relationships between \ltlf{} formulas.
During learning, temporal-causal knowledge expressed in TL-CDs is used to predict whether further exploration within a given episode is necessary and, if not, to short-circuit the exploration process.
Unfortunately, this method lacks rigorous theoretical guarantees.
Corazza et al.~\cite{pmlr-v236-corazza24a} extend this line of work with a provably correct method that directly integrates temporal-causal knowledge from TL-CDs into the reward machine.
Ultimately, both methods expedite convergence to optimal policies by exploiting knowledge of temporal causality---i.e., the long-term evolution of high-level events within the MDP.
In Section~\ref{sec:tlcds}, we provide a brief overview of TL-CDs and their integration with reward machines.

Our paper connects these two lines of work (decentralized MARL and temporal-causal RL).
In Section~\ref{sec:method}, we propose a method to exploit temporal-causal knowledge about the task at hand in order to (1) relax the theoretical compatibility criteria of DQPRM, broadening the scope of multi-agent tasks that can be decomposed and learned in a decentralized manner; and
(2) short-circuit certain explorations during decentralized learning, improving the sample efficiency of the learning process even further.
Moreover, we prove that our relaxed compatibility criterion is consistent with the original one, in the sense that passing the original criterion implies passing the relaxed one.
In Section~\ref{sec:case-studies}, we empirically demonstrate our method's effectiveness on the \emph{Generator Task} and introduce the \emph{Laboratory Task}, presenting experimental results for both.
An analysis of a third case study, the \emph{Buttons Task}, is provided in Appendix~\ref{sec:case_study_2}.

\section{Preliminaries}
\label{sec:preliminaries}

Aside from acting as a task specification, reward machines provide a form of finite memory for a team or an agent, tracking progress through the task.
We assume that during the agents' interaction with the MDP, relevant high-level events are labeled.
Reward machines use these event labels as inputs to transition between states and determine the appropriate rewards.
By modeling the temporal dependencies within the task, RMs guide agents through complex environments, especially in situations where rewards are sparse.
We first introduce the general definitions of Event-based Reward Machines (RMs) and RM-based Markov Decision Processes (RM-MDPs), which provide a foundational framework for handling task specification and reward functions.
In Section~\ref{sec:dmarl}, we will extend this formalism to the multi-agent setting, where decentralized coordination becomes crucial.

\begin{definition}[Event-based Reward Machine]
    \label{def:rm}
    An event-based RM $\rewardMachine{R} = \langle \rmStates, \rmInit, \marlEventSet, \rmTrans, \rmOut, \rmAcc \rangle$ is a tuple where
    $\rmStates$ is a finite set of states
    with an initial state $\rmInit \in \rmStates$,
    $\marlEventSet$ is the set of events,
    $\rmTrans : \rmStates \times \marlEventSet \to \rmStates$ is a (partial) transition function,
    $\rmOut : \rmStates \times \rmStates \to \reals$ is a (partial) function mapping transitions to rewards, and
    $\rmAcc \subseteq \rmStates$ is a set of terminal states that signal the end of the interaction.
\end{definition}

When the RM is in state $\rmCommonState \in \rmStates$ and reads an event $\marlAtomicEvent \in \marlEventSet$ such that $(\rmCommonState, \marlAtomicEvent) \in \text{Dom}(\rmTrans)$, it transitions into a new state $\rmCommonState' = \rmTrans(\rmCommonState, \marlAtomicEvent)$ and outputs a reward $\rmOut(\rmCommonState, \rmCommonState')$.
Otherwise, the RM remains in the same state and outputs a reward of $0$.
For finite event sequences, we define $\rmTrans(\rmCommonState, \epsilon) = \rmCommonState$ and $\rmTrans(\rmCommonState, \marlEventSeq \marlAtomicEvent) = \rmTrans(\rmTrans(\rmCommonState, \marlEventSeq), \marlAtomicEvent)$, where $\epsilon$ is the empty sequence, $\marlEventSeq \in (\marlEventSet)^*$, and $\marlAtomicEvent \in \marlEventSet$.
Note that $\rmTrans$ and $\rmOut$ are partial functions, i.e., we do not require the transition function to be defined for every input event from every state.
The term \emph{event-based RM} is a deliberate departure from prior work~\cite{neary_reward_2021}, which conflated event-driven RMs with propositional RMs under the generic ``RM'' label.
We focus on \emph{task completion} RMs which output a reward of $1$ upon reaching a terminal state $\rmCommonState \in \rmAcc$, and $0$ otherwise.
For brevity, we refer to task completion event-based reward machines simply as reward machines throughout the remainder of this work.

To connect an RM with the MDP (with set of states $\mdpStates$), we use a labeling function $\mdpLabel \colon \mdpStates \times \rmStates \to 2^{\marlEventSet}$.
The codomain of $\mdpLabel$ is $\marlEventAlphabet$ to model the possibility that certain events may occur concurrently.
We restrict the labeling function so that only one event may occur per step, per agent.\footnote{We explain this assumption in Section~\ref{sec:dmarl}, where we provide background for decentralized MARL.}
Because of this restriction and the compatibility criterion defined in Section~\ref{sec:dmarl}, the order an RM reads simultaneous events does not matter.
In other words, if an RM reads an event sequence $\marlEventSeq \in \text{Perm}(\mdpLabel(\mdpCommonState, \rmCommonState, \mdpCommonState'))$, the resulting state $\rmCommonState' = \rmTrans(\rmCommonState, \marlEventSeq)$ is well-defined (independent of the permutation of $\marlEventSeq$).

\begin{definition}[RM-MDP]
    \label{def:rm-mdp}
    A Reward Machine-based Markov Decision Process is a tuple $\mdp{M} = \langle \mdpStates, \mdpInit, \mdpActions, \mdpTrans, \mdpDiscount, \rewardMachine{R}, \mdpLabel \rangle$
    consisting of a finite state space $\mdpStates$,
    an initial state $\mdpInit \in \mdpStates$,
    a finite set of actions $\mdpActions$,
    a probabilistic transition function $\mdpTrans \colon \mdpStates \times \rmStates \times \mdpActions \times \mdpStates \to [0, 1]$,
    a discount factor $\mdpDiscount \in (0, 1]$,
    a reward machine $\rewardMachine{R}$ which captures the reward function,
    and a labeling function $\mdpLabel \colon \mdpStates \times \rmStates \times \mdpStates \to 2^{\marlEventSet}$.
    Note that the transition probabilities $\mdpTrans(\mdpCommonState' \mid \mdpCommonState, \rmCommonState, \mdpCommonAction)$ are conditioned on the state of the reward machine $\rmCommonState \in \rmStates$.
\end{definition}

Intuitively, if the MDP and the RM are in a joint state $(\mdpCommonState, \rmCommonState) \in \mdpStates \times \rmStates$, and the agent chooses an action $\mdpCommonAction \in \mdpActions$, the MDP transitions to a new state $\mdpCommonState' \sim \mdpTrans(\mdpCommonState, \rmCommonState, \mdpCommonAction)$.
The labeling function then outputs a set of simultaneous events $\mdpLabel(\mdpCommonState, \rmCommonState, \mdpCommonState')$, which the RM reads in any order, and transitions into $\rmCommonState' = \rmTrans(\rmCommonState, \mdpLabel(\mdpCommonState, \rmCommonState, \mdpCommonState'))$.

We say that a trajectory $\rmMdpTrajectory{n}$ is attainable within an RM-MDP $\mdp{M} = \langle \mdpStates, \mdpInit, \mdpActions, \mdpTrans, \mdpDiscount, \rewardMachine{R}, \mdpLabel \rangle$ when $(\mdpCommonState_0, \rmCommonState_0) = (\mdpInit, \rmInit)$,
and for all $i = 0, 1, \ldots, \allowbreak n-1$ we have
$\mdpTrans(\mdpCommonState_i, \rmCommonState_i, \mdpCommonAction_i, \mdpCommonState_{i+1}) > 0$ and $\rmCommonState_{i+1} = \rmTrans(\rmCommonState_i, \mdpLabel(\mdpCommonState_i, \rmCommonState_i, \mdpCommonState_{i+1}))$.
An event sequence $\marlEventSeq = \marlEventSeqs{n}$ is attainable if it can be obtained using a finite number of concatenation and swap operations from the label sequence $\propInput_i = \mdpLabel(\mdpCommonState_i, \rmCommonState_i, \mdpCommonState_{i+1})$ of an attainable trace.


The reward function of an RM-MDP is induced by the reward machine $\rewardMachine{R}$ via $\mdpReward((\mdpCommonState, \rmCommonState), \mdpCommonAction, (\mdpCommonState', \rmCommonState')) = \rmOut(\rmCommonState, \rmCommonState')$.
Because $\mdpReward$ is Markovian over the product space $\mdpStates \times \rmStates$, one may use Q-learning to find an optimal policy in an RM-MDP.

\section{Decentralized Multi-Agent Reinforcement Learning}
\label{sec:dmarl}

In this section, we outline the baseline DQPRM algorithm proposed by Neary et al.~\cite{neary_reward_2021}, including its assumptions and theoretical guarantees.
In decentralized MARL with reward machines, the overall team task is specified by an RM $\rewardMachine{R}$ defined over a global set of events $\marlEventSet$ (all events that may occur in the MDP).
However, not every agent can observe all events from $\marlEventSet$.
In order to capture this limitation, we define a local event set $\marlEventSet_i$ for each agent $i=1, \ldots, N$, where $N$ is the number of agents.
We assume that $\marlEventSet = \bigcup_{i=1}^{N} \marlEventSet_i$.

In the \emph{Generator Task} from Figure~\ref{fig:case-study-3}, the local event set of agent 1 is $\marlEventSet_1 = \mathset{\textPipe, \textDoor}$, modeling the capabilities to fix the pipe (observe \iconPipe{}) and open the door (observe \iconDoor{}).
The local event set of agent 2 is $\marlEventSet_2 = \mathset{\textDoor, \textGenerator}$, modeling the capabilities to observe the door being unlocked (\iconDoor{}) and power the generator (\iconGenerator{}).
As demonstrated by the event \textDoor{}, local event sets may overlap, i.e., there may be events that can be sensed by more than one agent.
We call such events \emph{shared events}, and write $\marlShared_{\marlAtomicEvent} = \mathset{i \mid \marlAtomicEvent \in \marlEventSet_i}$ for the set of agents that share event $\marlAtomicEvent$.
When $\vert \marlShared_{\marlAtomicEvent} \vert > 1$ for some $\marlAtomicEvent \in \marlEventSet$, a synchronization mechanism ought to be simulated during decentralized training episodes.
For example, since the event $\textDoor$ is under the control of agent 1, during decentralized training of agent 2, we simulate the communication of this event with probability $p > 0$ at every time-step.

In decentralized MARL, we aim to train each agent independently, so as to avoid learning a centralized policy.
The primary benefit of this approach is sample efficiency, as centralized policies induce a state space that grows exponentially with the number of agents.
Other benefits include privacy (limiting shared knowledge), and the fact that centralized policies may be infeasible due to practical concerns (differing capabilities, limited communication).

To facilitate decentralized training, we first need to capture the individual contribution of each agent towards the team goal in terms of the agent's local event set.
This can be achieved with the notion of \emph{Projected} Reward Machines, which we will now introduce.
We also refer to projected reward machines as \emph{local} RMs, to highlight that the projection is defined in terms of the local event set.

To define the projection of the team RM $\rewardMachine{R} = \langle \rmStates, \rmInit, \marlEventSet, \rmTrans, \rmOut, \rmAcc \rangle$ along a local event set $\marlEventSet_i \subseteq \marlEventSet$, we first introduce an equivalence relation $\marlIndistinguishability_i$ over states of $\rewardMachine{R}$.
Informally, two states $\rmCommonState_1, \rmCommonState_2 \in \rmStates$ are $\marlIndistinguishability_i$-equivalent if agent $i$ can not distinguish between them based on events in $\marlEventSet_i$.
Formally, $\marlIndistinguishability_i$ is the smallest equivalence relation such that for all $\rmCommonState_1, \rmCommonState_2, \rmCommonState_1', \rmCommonState_2' \in \rmStates$ we have (1) $\rmCommonState_1 \marlIndistinguishability_i \rmCommonState_2$ if there exists an event $e \in \marlEventSet \setminus \marlEventSet_i$ such that $\rmCommonState_2 = \rmTrans(\rmCommonState_1, e)$; and (2) if $\rmCommonState_1 \marlIndistinguishability_i \rmCommonState_1'$ and there exists an event $e \in \marlEventSet_i$ such that $\rmTrans(\rmCommonState_1, e) = \rmCommonState_2$ and $\rmTrans(\rmCommonState_1', e) = \rmCommonState_2'$, then $\rmCommonState_2 \marlIndistinguishability_i \rmCommonState_2'$.
We denote the set of equivalence classes of $\marlIndistinguishability_i$ with $\rmStates / \marlIndistinguishability_i$, and we will write $[\rmCommonState]_i$ for the $\marlIndistinguishability_i$-equivalence class of $\rmCommonState \in \rmStates$.

\begin{figure}[ht]
    \centering
    \begin{subfigure}[t]{.48\columnwidth}
        \centering
        \scalebox{0.8}{\begin{tikzpicture}[>=stealth, node distance=1.5cm, every state/.append style={scale=0.7}, initial text=]
    \node[state, initial] (0) at (0, 0) {$\rmInit^{(1)}$};
    \node[state] (1) at (1, 1) {$\rmCommonState_1^{(1)}$};
    \node[state] (2) at (1, -1) {$\rmCommonState_2^{(1)}$};
    \node[state, accepting] (3) at (2, 0) {$\rmCommonState_3^{(1)}$};

    \path[->] (0) edge node[above left] {\textPipe} (1);
    \path[->] (0) edge node[below left] {\textDoor} (2);
    \path[->] (1) edge node[above right] {\textDoor} (3);
    \path[->] (2) edge node[below right] {\textPipe} (3);
\end{tikzpicture}}
        \caption{$\marlEventSet_1 = \mathset{\textPipe, \textDoor}$}
        \label{fig:case-study-3-projection-1}
    \end{subfigure}%
    \hfill
    \begin{subfigure}[t]{.48\columnwidth}
        \centering
        \scalebox{0.8}{\begin{tikzpicture}[>=stealth, node distance=1.5cm, every state/.append style={scale=0.7}, initial text=]
    \node[state, initial] (0) at (0, 0) {$\rmInit^{(2)}$};
    \node[state] (1) at (1.5, 0) {$\rmCommonState_1^{(2)}$};
    \node[state, accepting] (2) at (3, 0) {$\rmCommonState_2^{(2)}$};

    \path[->] (0) edge node[above] {\textDoor} (1);
    \path[->] (1) edge node[above] {\textGenerator} (2);
\end{tikzpicture}}
        \caption{$\marlEventSet_2 = \mathset{\textDoor, \textGenerator}$}
        \label{fig:case-study-3-projection-2}
    \end{subfigure}
    \caption{Projections of the RM from Figure~\ref{fig:case-study-3-team-rm} along local event sets of agents 1 (Left) and 2 (Right).}
    \label{fig:case-study-3-projections}
\end{figure}
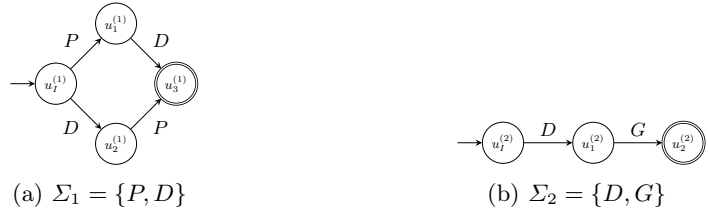

Condition (1) ensures that two states of $\rewardMachine{R}$ are indistinguishable for agent $i$ if a transition between them is triggered by an event outside of $\marlEventSet_i$.
Condition (2) (congruence) ensures that an event $e \in \marlEventSet_i$ may only trigger transitions to a unique successor equivalence class.
An algorithm to compute this equivalence relation with runtime $O(|\rmStates|^7 |\marlEventSet_i|^2 + |\rmStates|^5 |\marlEventSet_i| + |\rmStates|^4)$ is provided in appendix C of Wong~\cite{Wong1998OnTC}.
Using $\marlIndistinguishability_i$, we formalize the notion of a projected RM in Definition~\ref{def:projection}.
Figure~\ref{fig:case-study-3-projections} depicts the projections of the team RM from the \emph{Generator Task} along the local event sets of agents $1$ and $2$.

\begin{definition}[Projected Reward Machine]
    Given a reward machine $\rewardMachine{R} = \langle \rmStates, \rmInit, \marlEventSet, \rmTrans, \rmOut, \rmAcc \rangle$ and local event set $\marlEventSet_i$, the projection of $\rewardMachine{R}$ along $\marlEventSet_i$ is the RM
    $\rewardMachine{R}_i = (\rmStates_i, \rmInit^i, \rmTrans_i, \marlEventSet_i, \rmAcc_i)$ where
    $\rmStates_i = \rmStates / \marlIndistinguishability_i$,
    $\rmInit^i = [\rmInit]_i$,
    $\rmTrans_i : \rmStates_i \times \marlEventSet_i \to \rmStates_i$ is defined as $\rmCommonState^i_2 = \rmTrans_i(\rmCommonState^i_1, \marlAtomicEvent)$ for $\marlAtomicEvent \in \marlEventSet_i$ if and only if there exist $\rmCommonState_1, \rmCommonState_2 \in \rmStates$ such that $[\rmCommonState_1] = u^i_1$, $[\rmCommonState_2] = u^i_2$ and $\rmCommonState_2 = \rmTrans(\rmCommonState_1, \marlAtomicEvent)$ (and undefined otherwise),
    $\rmAcc_i = \mathset{\rmCommonState^i \in \rmStates_i \mid \exists \rmCommonState \in \rmAcc, \rmCommonState^i = [\rmCommonState]_i }$,
    and $\rmOut_i(\rmCommonState^i_1, \rmCommonState^i_2) = \indicatorNonset{\rmCommonState^i_2 \in \rmAcc_i \land \rmCommonState^i_1 \not\in \rmAcc_i}$.
    \label{def:projection}
\end{definition}

We model a multi-agent system with $N$ agents as an RM-MDP $\mdp{M} = \langle \mdpStates, \mdpInit, \mdpActions^N, \mdpTrans, \mdpDiscount, \rewardMachine{R}, \mdpLabel \rangle$ where the state space $\mdpStates$ is a Cartesian product of local state spaces, $\mdpStates = \mdpStates_1 \times \cdots \times \mdpStates_N$, the action space $\mdpActions^N$ is a product of individual action spaces (the same for each agent) $\mdpActions \times \cdots \times \mdpActions$, and the transition function $\mdpTrans : \mdpStates \times \rmStates \times \mdpActions^N \times \mdpStates$ can be decomposed as a product $\mdpTrans(\mdpCommonState, \rmCommonState, \mdpCommonAction, \mdpCommonState') = \prod_{i=1}^{N} \mdpTrans_i(\mdpCommonState_i, \rmCommonState^i, \mdpCommonAction_i, \mdpCommonState_i')$, where $\rmCommonState^i = [\rmCommonState]_i$ is the projection of the team state from $\rewardMachine{R}$ to the local RM state of agent $i$.
The dynamics $\mdpTrans_i$ of agent $i$ depend only on the agent's individual state and action, $(\mdpCommonState_i, \mdpCommonAction_i)$, and the events that the agent can sense (i.e., the state of its local RM $\rewardMachine{R}_i$).

To fully capture the local dynamics of each agent, DQPRM relies on the notion of local labeling functions $\mdpLabel_i : \mdpStates_i \times \rmStates_i \times \mdpStates_i \to 2^{\marlEventSet}$, where $\mdpStates_i$ is the local state space of agent $i$, and $\rmStates_i = \rmStates / \marlIndistinguishability_{i}$ are the states of its local RM.
Intuitively, $\mdpLabel_i(\mdpCommonState_i, \rmCommonState^i, \mdpCommonState_i')$ outputs events that \emph{could} be occurring from the perspective of agent $i$, given that the rest of the team state is unknown.
We note that local labeling functions must output at most a single event $\mathset{\marlAtomicEvent}$ (which corresponds to the assumption that only one event may occur per agent per time step), and that $\mdpLabel(\mdpCommonState, \rmCommonState, \mdpCommonState')$ outputs event $\marlAtomicEvent$ if and only if $\mdpLabel_i(\mdpCommonState_i, \rmCommonState^i, \mdpCommonState_i')$ output $\marlAtomicEvent$, for all $i \in \marlShared_\marlAtomicEvent$.
We call such label functions decomposable with respect to $\marlEventSet_1, \ldots, \marlEventSet_N$ (or just decomposable if the family $\mathset{\marlEventSet_i}_{i=1}^N$ is clear from the context).
\footnote{For more details on constructing local labeling functions see Neary et al.~\cite{neary_reward_2021}}
We define the projection of a finite event sequence $\marlEventSeq \in \marlEventSet^*$ onto the local event set $\marlEventSet_i$ as $\marlProjection_i(\marlEventSeq' \mdpCommonLabel) = \marlProjection_i(\marlEventSeq') \marlAtomicEvent$ if $\marlEventSeq = \marlEventSeq' \marlAtomicEvent$ and $\marlAtomicEvent \in \marlEventSet_i$, and $\marlProjection_i(\marlEventSeq \marlAtomicEvent) = \marlProjection_i(\marlEventSeq)$ otherwise (for the empty word $\epsilon$, $\marlProjection_i(\epsilon) = \epsilon$ for all $i = 1, \ldots, N$).

A key requirement for the task decomposition to be successful is that policies induced by projected RMs must combine to form a policy that satisfies the team RM.
This requirement is satisfied if the team RM is bisimilar to the parallel composition of its projections, which we refer to as the \emph{strict decomposition criterion}.
Intuitively, two reward machines $\rewardMachine{R}$ and $\rewardMachine{P}$ are bisimilar ($\rewardMachine{R} \bisim \rewardMachine{P}$) if there exists a relation between their states such that the transition behavior of related states is the same (initial and final states must be related).
The parallel composition of two RMs $\rewardMachine{R}_1$ and $\rewardMachine{R}_2$ is an RM $\rewardMachine{R}_1 \parallelComposition \rewardMachine{R}_2$ whose states are the Cartesian product of states of $\rewardMachine{R}$ and $\rewardMachine{P}$, initial (terminal) states are pairs of initial (terminal) states, and transitions are defined component-wise.\footnote{Formal definitions in Appendix~\ref{sec:definitions}. Visual explanation in Appendix~\ref{subsec:proof-relaxed-criterion}.}
Using the notions of bisimilarity and parallel composition of RMs, we can formalize the strict decomposition criterion for the team RM $\rewardMachine{R}$ and local RMs $\rewardMachine{R}_1, \ldots, \rewardMachine{R}_N$.
The criterion holds if $\rewardMachine{R} \bisim \parallelComposition_{i=1}^N \rewardMachine{R}_i$.
In that case, decentralized training of local policies will yield a combined policy with guaranteed bounds on the probability of solving the team task.
By definition, if local RMs $\rewardMachine{R}_i$ output $1$ (for all $i = 1, \ldots, N$), then their parallel composition also outputs $1$.
And because their parallel composition is bisimilar to the team RM, the team task is satisfied as well.
The converse also holds: if the team RM outputs a reward, then each local RM will output a reward.
This is formalized in Theorem~\ref{thm:decomposition}~\cite{neary_reward_2021}. 

\begin{theorem}[Strict Decomposition Criterion]
    If RM $\rewardMachine{R}$ and projections $\rewardMachine{R}_1, \ldots, \rewardMachine{R}_N$ satisfy the strict decomposition criterion, then
    $\forall\xi \in \marlEventSet^*$, $\rewardMachine{R}(\xi) = 1$ if and only if $\rewardMachine{R}_i(\marlProjection_i(\xi)) = 1, \forall i=1,\ldots,N$.
    Here, $\marlProjection_i(\marlEventSeq)$ is the projection of $\marlEventSeq$ to $\marlEventSet_i$ defined earlier.
    \label{thm:decomposition}
\end{theorem}

We provide the pseudocode for DQPRM in Algorithm~\ref{alg:dqprm}.
In brief, DQPRM runs $N$ concurrent episodes, where each agent moves independently, receiving observations, events, and rewards decoupled from the state of the other $N-1$ agents.
When acting in a team setting during testing, agents must synchronize on shared events.
This is the only communication channel necessary to execute single-agent policies, found by DQPRM, in a multi-agent RM-MDP.
When the local labeling function $\mdpLabel_i$ outputs a shared event $\marlAtomicEvent$, synchronization ensures that $\rewardMachine{R}_i$ reads $\marlAtomicEvent$ iff. $\mdpLabel_j$ outputs the same event $\marlAtomicEvent$ for all $j \in \marlShared_\marlAtomicEvent$.
To account for this fact during decentralized training, DQPRM simulates the synchronization signal with probability $p > 0$ (Line~\ref{line:sync}).
We summarize the guarantees for team performance in Theorem~\ref{thm:strict-decomposition-viability}.

\begin{algorithm}[h!]
    \caption{Decentralized training with DQPRM}
    \label{alg:dqprm}
    \LinesNumbered
    \KwIn{Team RM $\rewardMachine{R}$, local event sets $\marlEventSet_1, \ldots, \marlEventSet_N$, local labeling functions $\mdpLabel_1, \ldots, \mdpLabel_N$}
    \KwOut{Policies $\policy_1, \ldots, \policy_N$}

    Project $\rewardMachine{R}$ along local event sets $\marlEventSet_i$ to obtain projected RMs $\rewardMachine{R}_1, \ldots, \rewardMachine{R}_N$\;
    \If{$\rewardMachine{R} \not\bisim \rewardMachine{R}_1 \parallelComposition \cdots \parallelComposition \rewardMachine{R}_N$}{
        Reject: the strict decomposition criterion does not hold\;
    }

    $\text{Initialize}(\policy_1, \ldots, \policy_N)$\;

    \For{$m = 1$ \KwTo \textrm{numEpisodes}}{
        \For{$i = 1$ \KwTo $N$}{
            $\mdpCommonState_i \gets \mdpInit^{i}$, $\rmCommonState_i \gets \rmInit^{i}$\;
        }

        \For{$t = 1$ \KwTo \textrm{numSteps}}{
            \For{$i = 1$ \KwTo $N$}{
                \If{$\textrm{TaskComplete}_i(\rmCommonState_i)$}{\textbf{continue}\;}

                $\mdpCommonAction_i \gets \text{Sample}(\policy_i(\mdpCommonState_i, \rmCommonState_i))$\;
                $\mdpCommonState' \gets \text{Step}_i(s_i, u_i, a_i)$, $\rmCommonState' \gets \rmCommonState_i$, $\mdpCommonReward_i \gets 0$\;

                \tcp{$\mdpLabel_i(\mdpCommonState_i, \rmCommonState_i, \mdpCommonState')$ is either $\emptyset$ or $\{e\} \in \Sigma_i$}
                \For{$\marlAtomicEvent \in \mdpLabel_i(\mdpCommonState_i, \rmCommonState_i, \mdpCommonState')$}{
                    \If{$|\marlShared_{\marlAtomicEvent}| = 1$ \textbf{or} $\textrm{Rand}(0, 1) \leq p$\label{line:sync}}{
                        $\rmCommonState' \gets \rmTrans(\rmCommonState_i, \marlAtomicEvent)$;
                        $\mdpCommonReward_i \gets \mdpCommonReward_i + \rmOut(\rmCommonState_i, \rmCommonState')$;
                        $\rmCommonState_i \gets \rmCommonState'$\;
                    }
                }

                PolicyUpdate($\policy_i, \mdpCommonState_i, \rmCommonState_i, \mdpCommonAction_i, \mdpCommonState', \mdpCommonReward_i$)\;
                $\mdpCommonState_i \gets \mdpCommonState'$\;
            }
        }
    }

    \Return{$(\policy_1, \ldots, \policy_N)$}\;
\end{algorithm}

\begin{theorem}[Decomposition Viability]
    Given a team RM $\rewardMachine{R}$, local event sets $\marlEventSet_1, \ldots, \marlEventSet_N$, a decomposable label function $\mdpLabel$, and local labeling functions $\mdpLabel_1, \ldots, \mdpLabel_N$, assume that agents synchronize on shared events and $\rewardMachine{R} \bisim \parallelComposition_{i=1}^N \rewardMachine{R}_i$.
    Then for all team trajectories $\mdpCommonState_0 \rmCommonState_0 \cdots \mdpCommonState_k \rmCommonState_k$ and local trajectories $\mathset{\mdpCommonState_0^i \rmCommonState_0^i \cdots \mdpCommonState_k^i \rmCommonState_k^i}_{i=1}^N$, we have $\rewardMachine{R}(\mdpLabel(\mdpCommonState_0 \rmCommonState_0 \cdots \mdpCommonState_k \rmCommonState_k)) = 1$ if and only if $\rewardMachine{R}_i(\mdpLabel_i(\mdpCommonState_0^i \rmCommonState_0^i \cdots \mdpCommonState_k^i \rmCommonState_k^i)) = 1$ for all $i = 1, \ldots, N$.
    Furthermore, let $\valueFunction^{\policy}(\mdpInit)$ denote the team success probability and $\valueFunction^{\policy}_i(\mdpInit^i)$ the success probability of agent $i$.
    Then $\max\mathset{0, \valueFunction^{\policy}_1(\mdpInit) + \cdots + \valueFunction^{\policy}_N(\mdpInit) - (N - 1)} \leq \valueFunction^{\policy}(\mdpInit) \leq \min\mathset{\valueFunction^{\policy}_1(\mdpInit), \ldots, \valueFunction^{\policy}(\mdpInit)}$.
    \label{thm:strict-decomposition-viability}
\end{theorem}

Formal proofs of Theorems~\ref{thm:decomposition} and \ref{thm:strict-decomposition-viability} can be found in Neary et al.~\cite{neary_reward_2021}.
Informally, Theorem~\ref{thm:strict-decomposition-viability} states that if a task specification respects the strict decomposition criterion, decentralized training will yield local agent policies which, when combined, satisfy the team task (depending on the probability that each agents performs its own part of the task).
This result depends critically on Theorem~\ref{thm:decomposition}, which guarantees a correspondence between solving local tasks and solving the team task, and the assumption that the agents synchronize on shared events.

\section{Modeling Temporal-Causal Knowledge with TL-CDs}
\label{sec:tlcds}

In this section, we lay the groundwork for modeling temporal-causal knowledge in RM-MDPs, and extend the notion of Temporal Logic-based Causal Diagrams (TL-CDs), proposed in Paliwal et al.~\cite{paliwal2023reinforcement}, to Event-based Reward Machines.
Linear temporal logic over finite sequences ($\ltlf$) is a formal reasoning system that can capture causal and temporal properties of event sequences and RM-MDPs.
Aside from Boolean operators like $\lnot$ and $\lor$, $\ltlf$ introduces temporal operators such as $\globallyOp \psi$ (true if and only if $\psi$ holds for every element in the sequence), $\nextOp \psi$ (true iff. $\psi$ holds for the next element of the sequence), and $\psi \untilOp \varphi$ (true iff. $\psi$ holds until $\varphi$ becomes true, and $\varphi$ is true in some element of the sequence).
We also rely on the weak until operator $\psi \weakUntilOp \varphi$
(true iff. $\psi$ holds until $\varphi$ becomes true, but $\varphi$ is not required to become true).

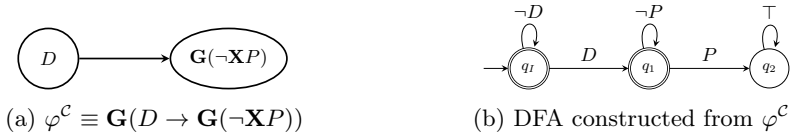
\begin{figure}[ht]
    \centering
    \begin{subfigure}[t]{0.48\columnwidth}
        \centering
        \scalebox{0.8}{\begin{tikzpicture}[>=stealth, node distance=1.5cm, every node/.style={thick, draw, ellipse, align=center, inner sep=1pt}]
    \node[minimum width=1cm, minimum height=1cm] (B1) {\textDoor};
    \node[minimum width=1cm, minimum height=1cm, right=of B1] (B2) {$\globallyOp(\lnot\nextOp\textPipe)$};

    \draw[->, thick] (B1) -- (B2);
\end{tikzpicture}}
        \caption{$\varphi^\causalDiagram \equiv \globallyOp(\textDoor \rightarrow \globallyOp(\lnot\nextOp\textPipe))$}
        \label{fig:case-study-3-tlcd-1}
    \end{subfigure}%
    \hfill
    \begin{subfigure}[t]{0.48\columnwidth}
        \centering
        \scalebox{0.8}{\begin{tikzpicture}[>=stealth, node distance=1.5cm, every state/.append style={scale=0.8}, initial text=]

    \node[state, initial, accepting] (0) at (0, 0) {$\dfaInitState$};
    \node[state, accepting] (1) at (2, 0) {$\dfaCommonState_1$};
    \node[state] (2) at (4, 0) {$\dfaCommonState_2$};

    \draw[->] (0) -- (1) node[midway, above] {$\textDoor$};
    \draw[->] (1) -- (2) node[midway, above] {$\textPipe$};

    \draw[->] (0) edge [loop above] node {$\lnot \textDoor$} ();
    \draw[->] (1) edge [loop above] node {$\lnot \textPipe$} ();
    \draw[->] (2) edge [loop above] node {$\top$} ();
\end{tikzpicture}}
        \caption{DFA constructed from $\varphi^\causalDiagram$}
        \label{fig:case-study-3-causal-dfa-1}
    \end{subfigure}
    \caption{TL-CD for the Generator Task (Left) and respective Causal DFA (Right)}
    \label{fig:case-study-3-tlcd-causal-dfa-1}
\end{figure}

TL-CDs are directed graphs where nodes are labeled with $\ltlf$ formulas and edges represent causal relationships between them, providing a structured notation to express temporal-causal knowledge.
To give the semantics of TL-CD $\causalDiagram$, one can construct an equivalent $\ltlf$ formula, $\varphi^\causalDiagram$, which captures the temporal-causal knowledge encoded in the graph.
The formula induced by the TL-CD in Figure~\ref{fig:case-study-3-tlcd-1}, $\globallyOp(\textDoor \rightarrow \globallyOp(\lnot\nextOp\textPipe))$, models sequences in which the event \textPipe{} never follows the event \textDoor{}.
In other words, it captures the effect of the one-way ramp (represented by blue arrows $\textcolor{blue}{\boldsymbol{\leftarrow}}$ in Figure~\ref{fig:case-study-3-gridworld}), which blocks agent 1 from returning to the region of the MDP containing the pipe (\iconPipe{}) after unlocking the door (\iconDoor{}).

To capture the semantics of a given TL-CD $\causalDiagram$ over event sequences in $\marlEventSet^*$, we construct an equivalent $\ltlf$ formula, $\Psi^{\causalDiagram}_{\marlEventSet}$, as the conjunction $\Psi^{\causalDiagram}_{\marlEventSet} \equiv \varphi^\causalDiagram \land \eventFormula$. The formula $\varphi^\causalDiagram$ encodes the temporal-causal knowledge expressed in $\causalDiagram$, and is defined as $\varphi^\causalDiagram \equiv \bigwedge_{\varphi \blacktriangleright \psi} \globallyOp(\varphi \rightarrow \psi) \label{eqn:tlcd}$, where $\varphi \blacktriangleright \psi$ iterates over edges that connect formulas $\varphi$ and $\psi$ in the TL-CD.
The formula $\eventFormula = \globallyOp ( \bigvee_{\marlAtomicEvent \in \marlEventSet} (\marlAtomicEvent_i \land (\bigwedge_{\marlAtomicEvent' \in \marlEventSet \setminus \mathset{\marlAtomicEvent}} \lnot \marlAtomicEvent')))$ restricts models of $\Psi^{\causalDiagram}_{\marlEventSet}$ to event sequences in $\marlEventSet^*$.
If $\Psi^\causalDiagram_\marlEventSet$ is true for an event sequence $\marlEventSeq = \marlEventSeqs{n}$, we will write $\marlEventSeq \models \Psi^\causalDiagram_\marlEventSet$.
We will say that a TL-CD $\causalDiagram$ holds for an MDP $\mdp{M}$ if for every event sequence $\marlEventSeq$ attainable in $\mdp{M}$, we have $\marlEventSeq \models \Psi^\causalDiagram_\marlEventSet$.
In order to simplify working with TL-CDs, we leverage the notion of deterministic finite automata (DFAs), formalized in Definition~\ref{def:dfa}.

\begin{definition}[Deterministic Finite Automaton]
    A DFA is a tuple $\dfa{C} = (\dfaStates, \dfaInitState, \marlEventSet, \dfaTrans, \dfaAcc)$ consisting of a finite set of states $\dfaStates$ with an initial state $\dfaInitState$, input alphabet $\marlEventSet$, deterministic transition function $\dfaTrans : \dfaStates \times \marlEventSet \to \dfaStates$, and a set of accepting states $\dfaAcc \subseteq \dfaStates$.
    \label{def:dfa}
\end{definition}


If the run of the DFA $\dfa{C}$ on an input string $\marlEventSeq$ ends in an accepting state $\dfaCommonState \in \dfaAcc$, we will write $\marlEventSeq \in \languageOf(\dfa{C})$.
Every TL-CD $\causalDiagram$ can be converted into an equivalent DFA $\dfa{C}$~\cite{paliwal2023reinforcement}, in the sense that for every $\marlEventSeq$, we have $\marlEventSeq \in \languageOf(\dfa{C}) \iff \marlEventSeq \models \Psi^\causalDiagram_\marlEventSet$.
We will refer to $\dfa{C}$ as the \emph{causal DFA}.
When illustrating causal DFAs, we will only consider the fragment constructed from $\varphi^\causalDiagram$, as the contribution of $\eventFormula$ is always the same and serves a technical purpose.
One can obtain the full causal DFA from the parallel composition of automata for $\varphi^\causalDiagram$ and $\eventFormula$.
In Figure~\ref{fig:case-study-3-causal-dfa-1}, we illustrate the DFA induced by the TL-CD which holds for the \emph{Generator Task}.

\section{Causal DQPRM}
\label{sec:method}


We enhance decentralized multi-agent RL by integrating temporal-causal knowledge through TL-CDs in two key ways.
\emph{First}, we enable valid decomposition of tasks rejected by DQPRM through causal constraints, preserving performance guarantees.
\emph{Second}, we accelerate learning by short-circuiting redundant exploration using TL-CD predictions, significantly improving sample efficiency in decentralized training.

\subsection{Relaxing the Decomposition Criterion in DQPRM}

The main assumption of the method proposed in Neary et al.~\cite{neary_reward_2021} is that the team RM is bisimilar to the parallel composition of the projected RMs.
As demonstrated by the decomposition in Figure~\ref{fig:case-study-3-projections}, this assumption is not always satisfied.
To illustrate this point, consider the event sequence $\marlEventSeq = \textDoor{} \textGenerator{} \textPipe{}$, which induces a reward of $1$ in both projections, but $0$ in the team RM.
Unfortunately, this means that Theorem~\ref{thm:strict-decomposition-viability} no longer guarantees a viable decomposition into local RMs.
As such, one can not expect DQPRM to solve the \emph{Generator Task}
(at least not before results analogous to Theorem~\ref{thm:decomposition} are shown to hold).

On the other hand, by constructing a strategy for each agent manually, one can see that decentralized learning ought to yield satisfactory results.
The reason for this lies in the fact that \iconPipe{} ($\textPipe$) and \iconDoor{} ($\textDoor$) are separated by a one-way ramp (Figure~\ref{fig:case-study-3-gridworld}).
Once agent 1 visits \iconDoor{}, then due to this one-way ramp, no attainable trajectory leads to \iconPipe{}.
In other words, while agent 1 satisfies its local task (Figure~\ref{fig:case-study-3-projection-1}) by observing $\textPipe$ and $\textDoor$ in any order, the structure of the MDP does not permit orders of the form
$\textPipe \rightarrow \cdots \rightarrow \textDoor$.
Therefore, Q-learning, equipped with knowledge of the current RM state of agent 1, ought to converge to a policy which avoids such paths.


As TL-CDs overapproximate the set of attainable event sequences within the MDP, they enable experts to encode temporal-causal knowledge—for instance, the fact that problematic sequences like $\marlEventSeq = \textDoor{} \textGenerator{} \textPipe{}$ are unattainable.
Given a team task RM $\rewardMachine{R}$ and local event sets $\marlEventSet_1, \ldots, \marlEventSet_N$ where $\rewardMachine{R} \not\bisim \parallelComposition_{i=1}^N \rewardMachine{R}_i$, our method leverages TL-CD $\causalDiagram$ to validate whether decomposition remains viable under the causal constraints of the RM-MDP.

To relax the strict bisimilarity assumption ($\rewardMachine{R} \bisim \parallelComposition_{i=1}^N \rewardMachine{R}_i$) from Theorem~\ref{thm:decomposition}, we introduce a modified bisimulation check:
$\rmCompDfa{R}{C} \bisim \rmCompDfa{P}{C}$,
where $\rewardMachine{P} = \parallelComposition_{i=1}^N \rewardMachine{R}_i$ represents the parallel composition of projected RMs, and $\rmCompDfa{R}{C}$, $\rmCompDfa{P}{C}$ denote their respective compositions with the causal DFA $\dfa{C}$ derived from $\causalDiagram$.

The parallel composition of an RM and a causal DFA synchronizes their transitions, meaning that $\rmCompDfa{R}{C}$ and $\rmCompDfa{P}{C}$ do not reward event sequences which do not satisfy the TL-CD $\causalDiagram$.
In Theorem~\ref{thm:decomposition-relaxed} we show that our relaxed decomposition criterion yields the same guarantees as the strict one, and in Theorem~\ref{thm:decomposition-decisions-agreement} we show that the two are compatible (provided $\causalDiagram$ holds for the RM-MDP).


As $\rewardMachine{R}$ and $\rewardMachine{P}$ are both deterministic finite automata, if $\rewardMachine{R} \not\bisim \rewardMachine{P}$, the issue is caused by event sequences $\marlEventSeq$ such that $\rewardMachine{R}(\marlEventSeq) \neq \rewardMachine{P}(\marlEventSeq)$~\cite{almeida_testing_2009}.
At its core, our method allows experts to exploit temporal-causal knowledge by finding a TL-CD $\causalDiagram$ such that
(1) $\causalDiagram$ holds for the RM-MDP, i.e., for every event sequence $\marlEventSeq$ attainable in $\mdp{M}$, we have $\marlEventSeq \models \Psi^\causalDiagram_\marlEventSet$; and
(2) $\rewardMachine{R}$, $\rewardMachine{P}$, and the causal DFA $\dfa{C}$ of the TL-CD $\causalDiagram$ pass the relaxed decomposition criterion, i.e. $\rmCompDfa{R}{C} \bisim \rmCompDfa{P}{C}$.

The first property ensures that the TL-CD overapproximates the RM-MDP and expresses correct causal knowledge about the environment.
The second property ensures that the causal knowledge added by the TL-CD is complete, in the sense that for every event sequence $\marlEventSeq$ such that $\rewardMachine{P}(\marlEventSeq) \neq \rewardMachine{R}(\marlEventSeq)$, we have $\marlEventSeq \not\models \Psi^\causalDiagram_\marlEventSet$ (i.e. $\marlEventSeq$ is not an attainable event sequence in the RM-MDP).
We use these properties to recover the important result that underpins team performance guarantees, and formalize our relaxed decomposition criterion in Theorem~\ref{thm:decomposition-relaxed}.
In Theorem~\ref{thm:decomposition-decisions-agreement}, we show that the relaxed and strict decomposition criteria are compatible, and in Theorem~\ref{thm:decomposition-viability} we provide a lower and upper bound on the team success probability, analogous to Theorem~\ref{thm:strict-decomposition-viability}.
The proofs for the theorems in this section are provided in Appendix~\ref{sec:proofs}.

\begin{theorem}[Relaxed Decomposition Criterion]
    Let $\rewardMachine{R}$ be the team task RM, and $\rewardMachine{R}_1, \ldots, \rewardMachine{R}_N$ be the projections of $\rewardMachine{R}$ onto the local event sets $\marlEventSet_1, \ldots, \marlEventSet_N$, respectively.
    If $\causalDiagram$ is a TL-CD with causal DFA $\dfa{C}$ such that $\rmCompDfa{R}{C} \bisim (\parallelComposition_{i=1}^N \rewardMachine{R}_i)\parallelComposition \dfa{C}$, then for all $\mdp{M}$-attainable event sequences $\marlEventSeq$, we have $\rewardMachine{R}(\marlEventSeq) = 1$ if and only if $\rewardMachine{R}_i(\marlProjection_i(\marlEventSeq)) = 1$ ($\forall i=1, \ldots, N$).
    Proof in Appendix~\ref{subsec:proof-decomposition-relaxed}.
    \label{thm:decomposition-relaxed}
\end{theorem}

Theorem~\ref{thm:decomposition-relaxed} states that if one can find a TL-CD $\causalDiagram$ which satisfies the two properties outlined above, then the team task RM can be decomposed into local task RMs such that on all attainable sequences, the team RM and the parallel composition of local RMs output the same reward.
Intuitively, if agents 1 through $N$ satisfy their local tasks (induce a reward of $1$ in projected RMs), then their combined behavior also induces a reward of $1$ in the parallel composition of projected RMs (which is bisimilar to the team RM).

If the strict decomposition criterion holds, it is not immediately clear that the relaxed criterion will also hold (given an arbitrary TL-CD that holds for the MDP).
In other words, Theorem~\ref{thm:decomposition} and Theorem~\ref{thm:decomposition-relaxed} present two different criteria for decomposing the team task RM (C1 and C2, respectively), and it is natural to ask whether these criteria are compatible.
Table~\ref{tab:decomposition-criteria-compatibility} summarizes the possible outcomes of comparing the two criteria.

\begin{table}[ht]
    \centering
    \begin{tabular}{@{}ccp{5cm}@{}}
        \toprule
        \textbf{C1} & \textbf{C2} & \textbf{Explanation}                   \\
        \midrule
        \faBan{}    & \faBan{}    & No decomposition (neither met)         \\
        \faBan{}    & \faCheck{}  & New decomposition (our method)         \\
        \faCheck{}  & \faBan{}    & \textit{Claim:} Impossible combination \\
        \faCheck{}  & \faCheck{}  & Known decomposition (both met)         \\
        \bottomrule
    \end{tabular}
    \caption{Results from Theorems~\ref{thm:decomposition-relaxed} \&~\ref{thm:decomposition-decisions-agreement}}
    \label{tab:decomposition-criteria-compatibility}
\end{table}

\begin{theorem}[Criterion Compatibility]
    Given a team task RM $\rewardMachine{R}$, a family of local event sets $\marlEventSet_1, \ldots, \marlEventSet_N$, and a TL-CD $\causalDiagram$ with causal DFA $\dfa{C}$, let $\rewardMachine{P} = \parallelComposition_{i=1}^N \rewardMachine{R}_i$ be the parallel composition of local task DFAs.
    Then $\rewardMachine{R} \bisim \rewardMachine{P} \Rightarrow \rmCompDfa{R}{C} \bisim \rmCompDfa{P}{C}$.
    In other words, the additional parallel composition with the causal DFA will not change the bisimulation decision if the original parallel composition is bisimilar to the team task DFA.
    Proof in Appendix~\ref{subsec:proof-relaxed-criterion}.
    \label{thm:decomposition-decisions-agreement}
\end{theorem}

In Theorem~\ref{thm:decomposition-decisions-agreement} we show that a parallel composition with an appropriate causal DFA introduced by our method will not make a task RM, that was originally decomposable, \emph{not} decomposable.
In other words, the relaxed criterion is a generalization of the original criterion, and so the two criteria are compatible.
Once the relationship between the two criteria has been established, the main result is given by Theorem~\ref{thm:decomposition-viability}, which provides lower and upper bounds on the probability of combined action success, analogous to Theorem~\ref{thm:strict-decomposition-viability}.
In order to derive the results, we also assume that agents synchronize on shared events in the team setting, and that the team labeling function $\mdpLabel$ is decomposable with respect to the local event sets with corresponding labeling functions $\mdpLabel_i$.



\begin{theorem}[Relaxed Decomposition Viability]
    Given a team RM $\rewardMachine{R}$, local event sets $\marlEventSet_1, \ldots, \marlEventSet_N$, a decomposable label function $\mdpLabel$, and local labeling functions $\mdpLabel_1, \ldots, \mdpLabel_N$, assume that agents synchronize on shared events and $\rmCompDfa{R}{C} \bisim (\parallelComposition_{i=1}^N \rewardMachine{R}_i)\parallelComposition \dfa{C}$.
    Then for all team trajectories $\mdpCommonState_0 \rmCommonState_0 \cdots \mdpCommonState_k \rmCommonState_k$ and local trajectories $\mathset{\mdpCommonState_0^i \rmCommonState_0^i \cdots \mdpCommonState_k^i \rmCommonState_k^i}_{i=1}^N$, we have $\rewardMachine{R}(\mdpLabel(\mdpCommonState_0 \rmCommonState_0 \cdots \mdpCommonState_k \rmCommonState_k)) = 1$ if and only if $\rewardMachine{R}_i(\mdpLabel_i(\mdpCommonState_0^i \rmCommonState_0^i \cdots \mdpCommonState_k^i \rmCommonState_k^i)) = 1$ for all $i = 1, \ldots, N$.
    Furthermore, we retain the bounds for the team success probability $\valueFunction^{\policy}(\mdpInit)$, i.e. $\max\mathset{0, \valueFunction^{\policy}_1(\mdpInit) + \cdots + \valueFunction^{\policy}_N(\mdpInit) - (N - 1)} \leq \valueFunction^{\policy}(\mdpInit) \leq \min\mathset{\valueFunction^{\policy}_1(\mdpInit), \ldots, \valueFunction^{\policy}(\mdpInit)}$.
    Proof in Appendix~\ref{subsec:proof-decomposition-viability}.
    \label{thm:decomposition-viability}
\end{theorem}

\subsection{Expediting RL with Temporal-Causal Knowledge}
\label{subsec:method-expediting-rl}


In the \emph{Generator Task} from Figure~\ref{fig:case-study-3}, if agent 1 unlocks the door \iconDoor{} before fixing the pipe \iconPipe{}, it will not be able to return and fix the pipe later, because its path is blocked by a one-way ramp.
Unfortunately, agent 1's projected reward machine, illustrated in Figure~\ref{fig:case-study-3-projection-1}, does not capture this information.
Due to this mismatch, during decentralized training episodes, agent 1 will tend to waste a large portion of time steps exploring trajectories which do not lead to a positive reward.

As discussed in Section~\ref{sec:tlcds}, one can use high-level symbolic knowledge in the form of the TL-CD on Figure~\ref{fig:case-study-3-tlcd-1} to capture this information.
While we first exploited this knowledge to check if the task specification for the \emph{Generator Task} is decomposable into local task specifications, we will now use it to expedite decentralized training for agent 1.
Note that the same TL-CD has been applied for both purposes: this is not always the case.
In our second case study, which covers the \emph{Laboratory Task}, we use a separate TL-CD to expedite decentralized training for two agents at once, but a different one to prove that the task specification passes the relaxed decomposition criterion.

Designing task specifications using reward machines is an error-prone and time-consuming task, and it is challenging to accommodate all possible scenarios and causal structures in advance.
Moreover, manually updating reward machines can lead to unintended consequences, ultimately inducing a different optimal policy.
Therefore, we investigate how to extend DQPRM to automatically incorporate temporal-causal knowledge about the environment, and help agents achieve a better balance exploration and exploitation, without adversely affecting performance in the original task.

To this end, we adapt the method proposed in Corazza et al.~\cite{pmlr-v236-corazza24a} to Event-based task-completion RMs.
The method relies on finding \emph{rejecting sink states} in the causal DFA $\dfa{C}$, i.e., states $\dfaCommonState_{\text{r.s.}} \in \dfaStatesOf{C} \setminus \dfaAccOf{C}$ such that $\dfaTrans_{\dfa{C}}(\dfaCommonState, \marlAtomicEvent) = \dfaCommonState$ for all $\marlAtomicEvent \in \marlEventSet_i$.
We implicitly consider causal DFAs to have at most one rejecting sink state $\dfaCommonState_{\text{r.s.}}$, and that an accepting state is reachable from all other states.
This can be achieved by minimization~\cite{sipser13}.
On Figure~\ref{fig:case-study-3-causal-dfa-1}, that is the state $\dfaCommonState_2$.
At its core, the method exploits the fact that once a label sequence $\marlEventSeq$ induces a transition into $\dfaCommonState_{\text{r.s.}}$, there is no continuation $\marlEventSeq \cdot \marlEventSeq'$ that will exit it, and are therefore, all such label sequences unattainable.

In order to identify explorations that will not be rewarded, the method computes $\tilde{\rewardMachine{R}}_i$, a modification of agent $i$'s projected RM $\rewardMachine{R}_i$ that induces the same optimal policy~\cite{pmlr-v236-corazza24a}, but embeds the temporal-causal knowledge captured by $\causalDiagram$.
The states of $\tilde{\rewardMachine{R}}_i$ are elements of $\rmStates^i \times \dfaStatesOf{C}$ (the Cartesian product of the states of $\rewardMachine{R}_i$ and the causal DFA of $\causalDiagram$), while the transition function $\tilde{\rmTrans_i}$ is defined as $\tilde{\rmTrans_i}((\rmCommonState, \dfaCommonState), \marlAtomicEvent) = (\rmTrans_i(\rmCommonState, \marlAtomicEvent), \dfaTrans_{\dfa{C}}(\dfaCommonState, \marlAtomicEvent))$.
The reward function $\tilde{\rmOut_i}$ is defined as $\tilde{\rmOut_i}((\rmCommonState, \dfaCommonState), (\rmCommonState', \dfaCommonState')) = \rmOut_i(\rmCommonState, \rmCommonState')$ if $\dfaCommonState \neq \dfaCommonState_{\text{r.s.}}$, and $-1$ otherwise.
\footnote{To justify this we note that, for task-completion RMs, $-1$ satisfies the reward bounds computed in Corazza et al.~\cite{pmlr-v236-corazza24a}.}

Because only unattainable label sequences (such as $\textDoor{} \textGenerator{} \textPipe{}$ in the \emph{Generator Task}) induce transitions into rejecting sink states, setting the reward to $-1$ when reaching such states does not affect the optimal policy.
The final step relies on computing the solution to the Bellman optimality equation over states of $\tilde{\rewardMachine{R}}_i$, $\valueFunction^{*}(\rmCommonState, \dfaCommonState) = \max_{\substack{\marlAtomicEvent \in \marlEventSet_i, \rmCommonState' = \tilde{\rmTrans}_i(\rmCommonState, \marlAtomicEvent)}} \left(\rmOut\left((\rmCommonState, \dfaCommonState), (\rmCommonState', \dfaCommonState')\right) + \valueFunction^{*}\left(\rmCommonState', \dfaCommonState'\right)\right)$.
In every step, the maximum possible return from the current episode is bounded from above by $\valueFunction^{*}(\rmCommonState, \dfaCommonState)$ and $0$ from below (by definition of task-completion RMs).
If $\tilde{\rewardMachine{R}}_i$ reaches a state $(\rmCommonState, \dfaCommonState)$ such that $\valueFunction^{*}(\rmCommonState, \dfaCommonState) = 0$ during decentralized training, then all policies induce the same return thereon, and the learning can stop.



We summarize our method in Algorithm~\ref{alg:causal-dqprm}, which we call Causal DQPRM.
The inputs for Causal DQPRM include a TL-CD $\causalDiagram$ which is used to ensure the task specification passes the relaxed decomposition criterion, along with a family of TL-CDs $\causalDiagram_1, \ldots, \causalDiagram_N$ (possibly a different one for each agent).
In Algorithm~\ref{alg:causal-dqprm}, $\dfaCommonState_i$ ranges over the states of causal DFA $\dfa{C}_i$, and $(\rmCommonState_i, \dfaCommonState_i)$ over the states of $\tilde{\rewardMachine{R}}_i$, the recomputed task specification for agent $i$.
In our experiments, we use $p=0.3$.

\begin{algorithm}[h]
    \caption{Decentralized training with Causal DQPRM}
    \label{alg:causal-dqprm}
    \LinesNumbered
    \SetKwInOut{Input}{Input}
    \SetKwInOut{Output}{Output}

    \Input{Team RM $\rewardMachine{R}$, local event sets $\marlEventSet_1, \ldots, \marlEventSet_N$, local labeling functions $\mdpLabel_1, \ldots, \mdpLabel_N$, TL-CDs $\causalDiagram, \causalDiagram_1, \ldots, \causalDiagram_N$}
    \Output{Policies $\policy_1, \ldots, \policy_N$}

    Project $\rewardMachine{R}$ along local event sets $\marlEventSet_i$ to obtain projected RMs $\rewardMachine{R}_1, \ldots, \rewardMachine{R}_N$\;
    \If{$\rmCompDfa{R}{C} \bisim (\parallelComposition_{i=1}^N \rewardMachine{R}_i)\parallelComposition \dfa{C}$}{
        Reject: the relaxed decomposition criterion does not hold\;
    }

    \For{$i = 1$ \KwTo $N$}{
        Compute $\tilde{\rewardMachine{R}}_i$ via value iteration over $\rewardMachine{R}_i\parallelComposition \dfa{C}_i$ \;
        $\mdpCommonState_i \gets \mdpInit^{i}$, $(\rmCommonState_i, \dfaCommonState_i) \gets (\rmInit^{i}, \dfaInitState^{i})$, $\textnormal{steps}_i \gets 0$\;
    }

    $\text{Initialize}(\policy_1, \ldots, \policy_N)$\;

    \For{$t = 1$ \KwTo $\textnormal{numEpisodes} * \textnormal{numSteps}$}{
    \For{$i = 1$ \KwTo $N$}{
    \If{$\textnormal{steps}_i > \textnormal{numSteps}$ \textbf{or} $\valueFunction^*_i(\rmCommonState_i, \dfaCommonState_i) = 0$}{
        $\mdpCommonState_i \gets \mdpInit^{i}$, $(\rmCommonState_i, \dfaCommonState_i) \gets (\rmInit^{i}, \dfaInitState^{i})$, $\textnormal{steps}_i \gets 0$\;
    }

    $\mdpCommonAction_i \gets \text{Sample}(\policy_i(\mdpCommonState_i, \rmCommonState_i))$\;
    $\mdpCommonState' \gets \text{Step}_i(s_i, u_i, a_i)$, $(\rmCommonState', \dfaCommonState') \gets (\rmCommonState_i, \dfaCommonState_i)$, $\mdpCommonReward_i \gets 0$\;

    \ForEach{$\marlAtomicEvent \in \mdpLabel_i(\mdpCommonState_i, \rmCommonState_i, \mdpCommonState')$}{
        \If{$|\marlShared_{\marlAtomicEvent}| = 1$ \textbf{or} $\textnormal{Rand}(0, 1) \leq p$}{
            $(\rmCommonState', \dfaCommonState') \gets \tilde{\rmTrans}((\rmCommonState_i, \dfaCommonState_i), \marlAtomicEvent)$\;
                $\mdpCommonReward_i \gets \mdpCommonReward_i + \tilde{\rmOut}((\rmCommonState_i, \dfaCommonState_i), \rmCommonState')$\;
                $(\rmCommonState_i, \dfaCommonState_i) \gets (\rmCommonState', \dfaCommonState')$\;
                }
                }

                PolicyUpdate($\policy_i, \mdpCommonState_i, \rmCommonState_i, \mdpCommonAction_i, \mdpCommonState', \mdpCommonReward_i$)\;
            $\mdpCommonState_i \gets \mdpCommonState'$, $\textnormal{steps}_i \gets \textnormal{steps}_i + 1$\;
        }
    }

    \Return $(\policy_1, \ldots, \policy_N)$\;
\end{algorithm}

\section{Case Studies}
\label{sec:case-studies}

We performed three case studies, validating our approach in the \emph{Generator Task}, described extensively throughout this paper, and two new domains: the \emph{Laboratory Task}, and the \emph{Buttons Task}.
This section will detail the results on the \emph{Generator Task} and the \emph{Laboratory Task}, while the \emph{Buttons Task} is described in Appendix~\ref{sec:case_study_2}.

\subsection{Case Study 1: Generator Task}
\label{sec:case_study_3}

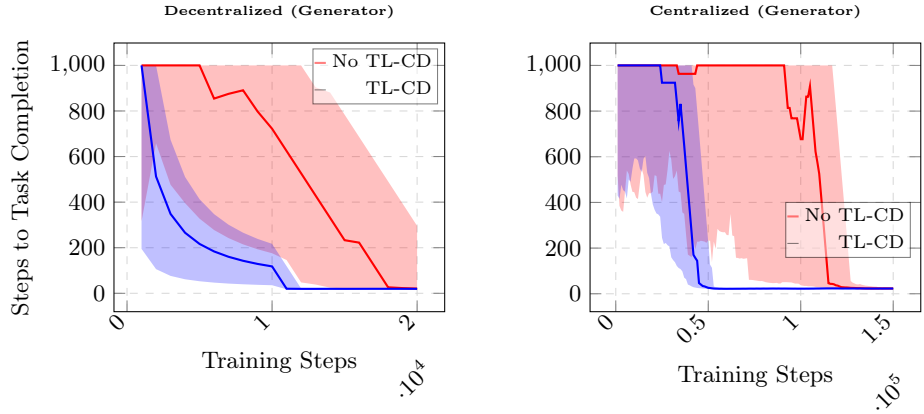
\begin{figure}[ht]
  \centering
  \begin{tikzpicture}[baseline]
    \begin{axis}[
        width=0.49\linewidth, 
        grid=major,
        grid style={dashed,gray!30},
        xlabel=Training Steps,
        ylabel=Steps to Task Completion,
        legend style={
            anchor=north east,
            legend cell align=right,
            draw=black,
            fill=white,
            text opacity=1,
            opacity=0.5,
            font=\scriptsize, 
            inner sep=0.2pt, 
            outer sep=0.2pt, 
            legend image post style={xscale=0.25}, 
          },
        x tick label style={rotate=45,anchor=east},
        tick label style={font=\small}, 
        label style={font=\small}, 
        ytick={0,200,400,600,800,1000},  
        title={\textbf{\tiny Decentralized (Generator)}},
      ]
      \addplot[
        thick,
        red
      ] table [y=prc_50, x=steps, col sep=comma] {results/final2/generator_DECENTRALIZED_NO_TLCD/2024-10-12_13-40-05.csv};
      \addlegendentry{No TL-CD}
      \addplot[
        name path=upper,
        draw=none
      ] table[y=prc_75, x=steps, col sep=comma] {results/final2/generator_DECENTRALIZED_NO_TLCD/2024-10-12_13-40-05.csv};
      \addplot[
        name path=lower,
        draw=none
      ] table[y=prc_25, x=steps, col sep=comma] {results/final2/generator_DECENTRALIZED_NO_TLCD/2024-10-12_13-40-05.csv};
      \addplot[
        fill=red,
        fill opacity=0.25
      ] fill between[of=upper and lower];
      \addplot[
        thick,
        blue,
        draw=blue
      ] table [y=prc_50, x=steps, col sep=comma] {results/final2/generator_DECENTRALIZED_TLCD/2024-10-12_13-42-32.csv};
      \addlegendentry{TL-CD}
      \addplot[
        name path=upper_tlcd,
        draw=none
      ] table[y=prc_75, x=steps, col sep=comma] {results/final2/generator_DECENTRALIZED_TLCD/2024-10-12_13-42-32.csv};
      \addplot[
        name path=lower_tlcd,
        draw=none
      ] table[y=prc_25, x=steps, col sep=comma] {results/final2/generator_DECENTRALIZED_TLCD/2024-10-12_13-42-32.csv};
      \addplot [
        fill=blue,
        fill opacity=0.25
      ] fill between[of=upper_tlcd and lower_tlcd];
    \end{axis}
  \end{tikzpicture}
  \hfill%
  \begin{tikzpicture}[baseline]
    \begin{axis}[
        width=0.49\linewidth, 
        grid=major,
        grid style={dashed,gray!30},
        xlabel=Training Steps,
        ylabel={}, 
        legend style={
            at={(0.97,0.20)}, 
            anchor=south east,
            legend cell align=right,
            draw=black,
            fill=white,
            text opacity=1,
            opacity=0.5,
            font=\scriptsize, 
            inner sep=0.2pt, 
            outer sep=0.2pt, 
            legend image post style={xscale=0.25}, 
          },
        x tick label style={rotate=45,anchor=east},
        tick label style={font=\small},
        label style={font=\small},
        ytick={0,200,400,600,800,1000},  
        title={\textbf{\tiny Centralized (Generator)}},
      ]
      \addplot[
        thick,
        red
      ] table [y=prc_50, x=steps, col sep=comma] {results/final2/generator_CENTRALIZED_NO_TLCD/2024-10-12_14-12-31.csv};
      \addlegendentry{No TL-CD}
      \addplot[
        name path=upper,
        draw=none
      ] table[y=prc_75, x=steps, col sep=comma] {results/final2/generator_CENTRALIZED_NO_TLCD/2024-10-12_14-12-31.csv};
      \addplot[
        name path=lower,
        draw=none
      ] table[y=prc_25, x=steps, col sep=comma] {results/final2/generator_CENTRALIZED_NO_TLCD/2024-10-12_14-12-31.csv};
      \addplot [
        fill=red,
        fill opacity=0.25
      ] fill between[of=upper and lower];
      \addplot[
        thick,
        blue
      ] table [y=prc_50, x=steps, col sep=comma] {results/final2/generator_CENTRALIZED_TLCD/2024-10-12_14-44-05.csv};
      \addlegendentry{TL-CD}
      \addplot[
        name path=upper_tlcd,
        draw=none
      ] table[y=prc_75, x=steps, col sep=comma] {results/final2/generator_CENTRALIZED_TLCD/2024-10-12_14-44-05.csv};
      \addplot[
        name path=lower_tlcd,
        draw=none
      ] table[y=prc_25, x=steps, col sep=comma] {results/final2/generator_CENTRALIZED_TLCD/2024-10-12_14-44-05.csv};
      \addplot [
        fill=blue,
        fill opacity=0.25
      ] fill between[of=upper_tlcd and lower_tlcd];
    \end{axis}
  \end{tikzpicture}
  \caption{\emph{Generator Task} study. Aggregated results from $50$ independent runs.}
  \label{fig:case-study-3-results}
\end{figure}

Our first case study is an ablation of our approach on the \emph{Generator Task}, with results shown in Figure~\ref{fig:case-study-3-results}.
We compare the average steps needed for task completion per training step.
The baseline we compare against is a centralized controller found with Q-learning, corresponding to the \emph{No TL-CD} plot on Figure~\ref{fig:case-study-3-results}, \emph{Centralized}.

Adding temporal-causal information improves team performance, even with centralized training.
Decentralized training, enabled by our relaxed decomposition check, is significantly more efficient, taking an order of magnitude fewer steps to converge.
In our two additional case studies, we perform the same analysis for the \emph{Laboratory Task} and the \emph{Buttons Task}.

\subsection{Case Study 2: Laboratory Task}
\label{sec:case_study_4}

\begin{figure}[ht]
  \centering
  \begin{tikzpicture}[baseline]
    \begin{axis}[
        width=0.49\linewidth, 
        grid=major,
        grid style={dashed,gray!30},
        xlabel=Training Steps,
        ylabel=Steps to Task Completion,
        legend style={
            anchor=north east,
            legend cell align=right,
            draw=black,
            fill=white,
            text opacity=1,
            opacity=0.5,
            font=\scriptsize, 
            inner sep=0.2pt, 
            outer sep=0.2pt, 
            legend image post style={xscale=0.25}, 
        },
        x tick label style={rotate=45,anchor=east},
        tick label style={font=\small}, 
        label style={font=\small}, 
        ytick={0,200,400,600,800,1000},  
        title={\textbf{\tiny Decentralized (Laboratory)}},
      ]
      \addplot[
        thick,
        red
      ] table [y=prc_50, x=steps, col sep=comma] {results/final2/laboratory_DECENTRALIZED_NO_TLCD/2024-10-11_22-46-11.csv};
      \addlegendentry{No TL-CD}
      \addplot[
        name path=upper,
        draw=none
      ] table[y=prc_75, x=steps, col sep=comma] {results/final2/laboratory_DECENTRALIZED_NO_TLCD/2024-10-11_22-46-11.csv};
      \addplot[
        name path=lower,
        draw=none
      ] table[y=prc_25, x=steps, col sep=comma] {results/final2/laboratory_DECENTRALIZED_NO_TLCD/2024-10-11_22-46-11.csv};
      \addplot[
        fill=red,
        fill opacity=0.25
      ] fill between[of=upper and lower];
      \addplot[
        thick,
        blue,
        draw=blue
      ] table [y=prc_50, x=steps, col sep=comma] {results/final2/laboratory_DECENTRALIZED_TLCD/2024-10-11_22-53-34.csv};
      \addlegendentry{TL-CD}
      \addplot[
        name path=upper_tlcd,
        draw=none
      ] table[y=prc_75, x=steps, col sep=comma] {results/final2/laboratory_DECENTRALIZED_TLCD/2024-10-11_22-53-34.csv};
      \addplot[
        name path=lower_tlcd,
        draw=none
      ] table[y=prc_25, x=steps, col sep=comma] {results/final2/laboratory_DECENTRALIZED_TLCD/2024-10-11_22-53-34.csv};
      \addplot [
        fill=blue,
        fill opacity=0.25
      ] fill between[of=upper_tlcd and lower_tlcd];
    \end{axis}
  \end{tikzpicture}
  \hfill%
  \begin{tikzpicture}[baseline]
    \begin{axis}[
        width=0.49\linewidth, 
        grid=major,
        grid style={dashed,gray!30},
        xlabel=Training Steps,
        ylabel={}, 
        legend style={
            at={(0.97,0.20)}, 
            anchor=south east,
            legend cell align=right,
            draw=black,
            fill=white,
            text opacity=1,
            opacity=0.5,
            font=\scriptsize, 
            inner sep=0.2pt, 
            outer sep=0.2pt, 
            legend image post style={xscale=0.25}, 
        },
        x tick label style={rotate=45,anchor=east},
        tick label style={font=\small},
        label style={font=\small},
        ytick={0,200,400,600,800,1000},  
        title={\textbf{\tiny Centralized (Laboratory)}},
      ]
      \addplot[
        thick,
        red
      ] table [y=prc_50, x=steps, col sep=comma] {results/final2/laboratory_CENTRALIZED_NO_TLCD/2024-10-12_04-42-12.csv};
      \addlegendentry{No TL-CD}
      \addplot[
        name path=upper,
        draw=none
      ] table[y=prc_75, x=steps, col sep=comma] {results/final2/laboratory_CENTRALIZED_NO_TLCD/2024-10-12_04-42-12.csv};
      \addplot[
        name path=lower,
        draw=none
      ] table[y=prc_25, x=steps, col sep=comma] {results/final2/laboratory_CENTRALIZED_NO_TLCD/2024-10-12_04-42-12.csv};
      \addplot [
        fill=red,
        fill opacity=0.25
      ] fill between[of=upper and lower];
      \addplot[
        thick,
        blue
      ] table [y=prc_50, x=steps, col sep=comma] {results/final2/laboratory_CENTRALIZED_TLCD/2024-10-12_09-42-54.csv};
      \addlegendentry{TL-CD}
      \addplot[
        name path=upper_tlcd,
        draw=none
      ] table[y=prc_75, x=steps, col sep=comma] {results/final2/laboratory_CENTRALIZED_TLCD/2024-10-12_09-42-54.csv};
      \addplot[
        name path=lower_tlcd,
        draw=none
      ] table[y=prc_25, x=steps, col sep=comma] {results/final2/laboratory_CENTRALIZED_TLCD/2024-10-12_09-42-54.csv};
      \addplot [
        fill=blue,
        fill opacity=0.25
      ] fill between[of=upper_tlcd and lower_tlcd];
    \end{axis}
  \end{tikzpicture}
  \caption{\emph{Laboratory Task} study. Aggregated results from $50$ independent runs.}
  \label{fig:case-study-4-results}
\end{figure}
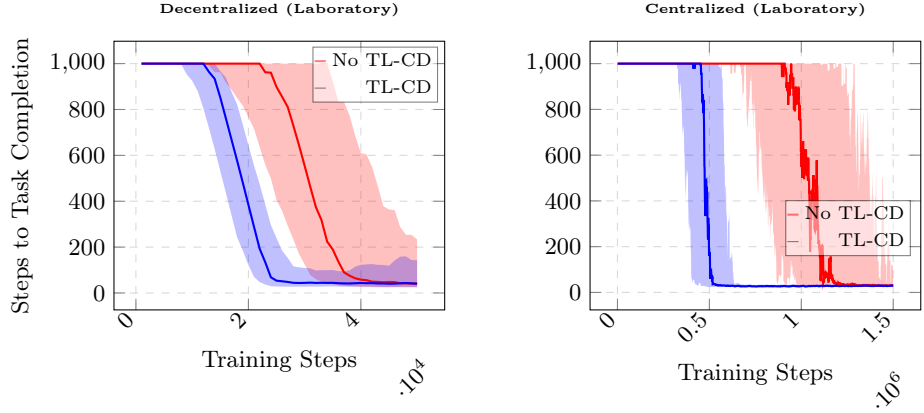








In the \emph{Laboratory Task}, two agents are tasked with aiding with an accident that occurred in a laboratory.
There are two possible types of accidents that may have occurred: a fire, represented by \faFire{}, or a radioactive spill, represented by \faRadiation{}.
Agent 1 is equipped with heat sensors, but not with radiation sensors, vice-versa for agent 2.

The two accident types are mutually exclusive.
Once the agents enter the conveyor belt, represented by \faCog, which leads them to the laboratory, either agent 1 will observe \faFire{}, or agent 2 will observe \faRadiation{} (with $50\%$ probability each).
Once inside, the agents must, depending on the type of the accident, converge to a tool which will provide aid (\faFireExtinguisher{} in case of fire, and \faBriefcaseMedical{} in case of radiation).

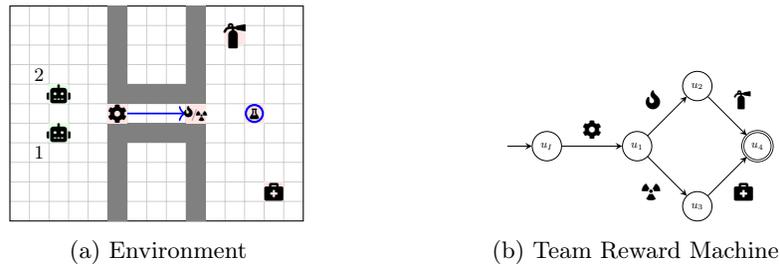
\begin{figure}[!htbp]
  \centering
  \begin{subfigure}{.48\columnwidth}
    \centering
    \scalebox{0.8}{\begin{tikzpicture}[scale=0.325]

\draw[opacity=0.2] (0,0) grid (15,11);
\draw[opacity=1] (0,0) rectangle (15,11);

\fill[black!50] (5,0) rectangle (6,4);

\fill[black!50] (5,7) rectangle (6,11);

\fill[black!50] (9,0) rectangle (10,4);

\fill[black!50] (9,7) rectangle (10,11);

\fill[black!50] (5,4) rectangle (10,5);

\fill[black!50] (5,6) rectangle (10,7);

\draw[->, blue, thick] (6, 5.5) -- (9,5.5);

\fill[green!10] (2,4) rectangle (3,5);
\node at (2.5,4.5) {\faRobot};
\node at (1.5,3.5) {1};

\fill[green!10] (2,6) rectangle (3,7);
\node at (2.5,6.5) {\faRobot};
\node at (1.5,7.5) {2};

\fill[red!10] (5,5) rectangle (6,6);
\node at (5.5,5.5) {\faCog};

\fill[red!10] (9,5) rectangle (10,6);
\node at (9.5,5.5) {\scalebox{0.6}{\raisebox{0.4ex}{\faFire{}}\kern-0.1em\slash\kern-0.1em\raisebox{-0.4ex}{\faRadiation{}}}};

\fill[red!10] (11,9) rectangle (12,10);
\node at (11.5,9.5) {\large \faFireExtinguisher};

\fill[red!10] (13,1) rectangle (14,2);
\node at (13.5,1.5) {\faBriefcaseMedical};

\fill[red!10] (12,5) rectangle (13,6);
\node at (12.5, 5.5) {\tikz[baseline=(char.base)]{\node[line width=1pt,blue,draw,circle,inner sep=0.5pt, text=black] (char) {\tiny \faFlask};}};

\end{tikzpicture}}
    \caption{Environment}
    \label{fig:case-study-4-gridworld}
  \end{subfigure}%
  \hfill
  \begin{subfigure}{.48\columnwidth}
    \centering
    \scalebox{0.8}{\begin{tikzpicture}[>=stealth, node distance=1.5cm, every state/.append style={scale=0.6}, initial text=]
    \node[state, initial] (0) at (0, 0) {$\rmInit$};
    \node[state] (1) at (1.5, 0) {$\rmCommonState_1$};
    \node[state] (2) at (2.5, 1) {$\rmCommonState_2$};
    \node[state] (3) at (2.5, -1) {$\rmCommonState_3$};
    \node[state, accepting] (4) at (3.5, 0) {$\rmCommonState_4$};

    \path[->] (0) edge node[above] {\faCog} (1);
    \path[->] (1) edge node[above left] {\faFire} (2);
    \path[->] (1) edge node[below left] {\faRadiation} (3);
    \path[->] (2) edge node[above right] {\faFireExtinguisher} (4);
    \path[->] (3) edge node[below right] {\faBriefcaseMedical} (4);
\end{tikzpicture}}
    \caption{Team Reward Machine}
    \label{fig:case-study-4-team-rm}
  \end{subfigure}
  \caption{\emph{Laboratory Task}}
  \label{fig:case-study-4}
\end{figure}

We illustrate the environment for the \emph{Laboratory Task} in Figure~\ref{fig:case-study-4-gridworld}, and the team reward machine which provides the task specification in Figure~\ref{fig:case-study-4-team-rm}.
As shown by the results in Figure~\ref{fig:case-study-4-results}, Causal DQPRM is able to solve this task, and leverages temporal-causal information to significantly increase sample efficiency compared to a centralized controller.

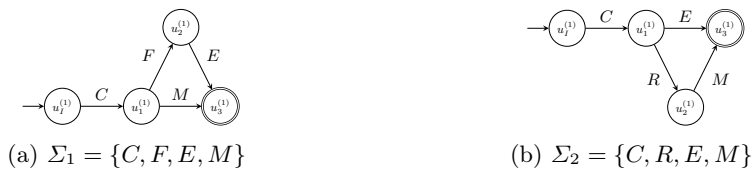
\begin{figure}[ht]
  \centering
  \begin{subfigure}[t]{.45\columnwidth}
    \centering
    \scalebox{0.7}{\begin{tikzpicture}[>=stealth, node distance=2.5cm, every state/.append style={scale=0.7}, initial text=]
    \node[state, initial] (0) at (0, 0) {$\rmInit^{(1)}$};
    \node[state] (1) at (1.5, 0) {$\rmCommonState_1^{(1)}$};
    \node[state] (2) at (2.25, 1.5) {$\rmCommonState_2^{(1)}$};
    \node[state, accepting] (3) at (3, 0) {$\rmCommonState_3^{(1)}$};

    \path[->] (0) edge node[above] {$C$} (1);
    \path[->] (1) edge node[above left] {$F$} (2);
    \path[->] (1) edge node[above] {$M$} (3);
    \path[->] (2) edge node[above right] {$E$} (3);
\end{tikzpicture}}
    \caption{$\marlEventSet_1 = \mathset{C, F, E, M}$}
    \label{fig:case-study-4-projection-1}
  \end{subfigure}%
  \hfill
  \begin{subfigure}[t]{.45\columnwidth}
    \centering
    \scalebox{0.7}{\begin{tikzpicture}[>=stealth, node distance=2.5cm, every state/.append style={scale=0.7}, initial text=]
    \node[state, initial] (0) at (0, 0) {$\rmInit^{(1)}$};
    \node[state] (1) at (1.5, 0) {$\rmCommonState_1^{(1)}$};
    \node[state] (2) at (2.25, -1.5) {$\rmCommonState_2^{(1)}$};
    \node[state, accepting] (3) at (3, 0) {$\rmCommonState_3^{(1)}$};

    \path[->] (0) edge node[above] {$C$} (1);
    \path[->] (1) edge node[below left] {$R$} (2);
    \path[->] (1) edge node[above] {$E$} (3);
    \path[->] (2) edge node[below right] {$M$} (3);
\end{tikzpicture}}
    \caption{$\marlEventSet_2 = \mathset{C, R, E, M}$}
    \label{fig:case-study-4-projection-2}
  \end{subfigure}
  \caption{Projections of the RM from Figure~\ref{fig:case-study-4-team-rm} along local event sets of agents 1 (Left) and 2 (Right).}
  \label{fig:case-study-4-projections}
\end{figure}

On Figure~\ref{fig:case-study-4-projections}, we illustrate the local reward machines of agents 1 and 2 in the \emph{Laboratory Task}.

Our experiments were conducted on a single machine with AMD Ryzen 7 5825U and 6.9G of RAM, except for the centralized case of the Buttons task, where we used a machine with AMD EPYC 7742 64-Core Processor and 515G of RAM, due to the extremely large state space.




\section{Conclusion}

We introduced a framework for integrating temporal-causal knowledge, formalized via TL-CDs, into decentralized multi-agent reinforcement learning, enabling both relaxed task decomposition and accelerated policy learning.
Experimentally validated across three case studies, our method improves the applicability of decentralized training guarantees while achieving higher success rates and sample efficiency compared to baseline approaches.
By automating the incorporation of expert knowledge into task specifications and learning processes, this work enables scalable solutions to complex, temporally structured multi-agent tasks.

\begin{credits}
      \subsubsection{\ackname}
      This work was supported in part by the National Science Foundation (NSF) under Grants CNS 2304863 and CNS 2339774, and in part by the Office of Naval Research (ONR) under Grant N00014-23-1-2505.

      \subsubsection{\discintname}
      The authors have no competing interests to declare that are relevant to the content of this article.
\end{credits}

\bibliographystyle{splncs04}
\bibliography{references}

\clearpage
\appendix
\section{Parallel Composition and Bisimilirity of Reward Machines}
\label{sec:definitions}

In this section, we introduce the formal definitions of parallel composition and bisimilarity for reward machines.

\begin{definition}[Parallel Composition of Reward Machines]
    Let $\rewardMachine{R}_1 = \langle \rmStates_1, \rmInit^1, \marlEventSet_1, \rmTrans_1, \rmOut_1, \rmAcc_1 \rangle$ and $\rewardMachine{R}_2 = \langle \rmStates_2, \rmInit^2, \marlEventSet_2, \rmTrans_2, \rmOut_2, \rmAcc_2 \rangle$ be two reward machines.
    The parallel composition of $\rewardMachine{R}_1$ and $\rewardMachine{R}_2$ is a new reward machine $\rewardMachine{P} = \langle \rmStates, \rmInit, \marlEventSet, \rmTrans, \rmOut, \rmAcc \rangle$ where the

    \begin{itemize}
        \item set of states is $\rmStates = \rmStates_1 \times \rmStates_2$,
        \item initial states is $\rmInit = (\rmInit^1, \rmInit^2)$,
        \item input alphabet is $\marlEventSet = \marlEventSet_1 \cup \marlEventSet_2$
        \item transition function is defined as \[
                  \rmTrans((\rmCommonState_1, \rmCommonState_2), \marlAtomicEvent) =
                  \begin{cases}
                      (\rmTrans_1(\rmCommonState_1, \marlAtomicEvent), \rmTrans_2(\rmCommonState_2, \marlAtomicEvent)) & \text{if } \rmTrans_1(\rmCommonState_1, \marlAtomicEvent) \text{ and } \\\rmTrans_2(\rmCommonState_2, \marlAtomicEvent) \text{ are defined} \\
                      (\rmTrans_1(\rmCommonState_1, \marlAtomicEvent), \rmCommonState_2)                               & \text{if only } \rmTrans_1(\rmCommonState_1, \marlAtomicEvent) \text{ is defined}                                                         \\
                      (\rmCommonState_1, \rmTrans_2(\rmCommonState_2, \marlAtomicEvent))                               & \text{if only } \rmTrans_2(\rmCommonState_2, \marlAtomicEvent) \text{ is defined}                                                         \\
                      \text{undefined}                                                                                 & \text{otherwise}
                  \end{cases}
              \]
        \item set of terminal states is $\rmAcc = \rmAcc_1 \times \rmAcc_2$; and
        \item output function $\rmOut : \rmStates \times \rmStates \to \reals$ is defined as $\rmOut(\rmCommonState, \rmCommonState') = 1$ if $\rmCommonState' \in \rmAcc$ and $\rmCommonState \not\in \rmAcc$ and $0$ otherwise.
    \end{itemize}

    \label{def:parallel-composition}
\end{definition}

\begin{definition}[Bisimilirity of Reward Machines]
    Let $\rewardMachine{R}_1 = \\ \langle \rmStates_1, \rmInit^1, \marlEventSet, \rmTrans_1, \rmOut_1, \rmAcc_1 \rangle$ and $\rewardMachine{R}_2 = \langle \rmStates_2, \rmInit^2, \marlEventSet, \rmTrans_2, \rmOut_2, \rmAcc_2 \rangle$ be two reward machines with the same event set $\marlEventSet$.
    $\rewardMachine{R}_1$ and $\rewardMachine{R}_2$ are bisimilar ($\rewardMachine{R}_1 \bisim \rewardMachine{R}_2$) if there exists a relation $\mathcal{R} \subseteq \rmStates_1 \times \rmStates_2$ such that

    \begin{enumerate}
        \item $(\rmInit^1, \rmInit^2) \in \mathcal{R}$.

        \item For every $(\rmCommonState_1, \rmCommonState_2) \in \mathcal{R}$ and $e \in \marlEventSet$:
              \begin{itemize}
                  \item \textbf{(Accepting)} $\rmCommonState_1 \in \rmAcc_1$ if and only if $\rmCommonState_2 \in \rmAcc_2$.

                  \item \textbf{(Forth)} if $\rmTrans_1(\rmCommonState_1, e) = \rmCommonState_1'$, then there exists $\rmCommonState_2' \in \rmStates_2$ such that $\rmTrans_2(\rmCommonState_2, e) = \rmCommonState_2'$ and $(\rmCommonState_1', \rmCommonState_2') \in \mathcal{R}$.

                  \item \textbf{(Back)} if $\rmTrans_2(\rmCommonState_2, e) = \rmCommonState_2'$, then there exists $\rmCommonState_1' \in \rmStates_1$ such that $\rmTrans_1(\rmCommonState_1, e) = \rmCommonState_1'$ and $(\rmCommonState_1', \rmCommonState_2') \in \mathcal{R}$.
              \end{itemize}
    \end{enumerate}

    \label{def:bisimilarity}
\end{definition}
\section{Proofs}
\label{sec:proofs}

\subsection{Proof of Theorem~\ref{thm:decomposition-relaxed} (Relaxed Decomposition Criterion)}
\label{subsec:proof-decomposition-relaxed}

\begin{proof}
    Let $\rewardMachine{R}$ be the team task RM, and $\rewardMachine{R}_1, \ldots, \rewardMachine{R}_N$ be the projections of $\rewardMachine{R}$ onto the local event sets $\marlEventSet_1, \ldots, \marlEventSet_N$, respectively.
    Let $\causalDiagram$ be a TL-CD that holds for $\mdp{M}$, with equivalent causal DFA $\dfa{C}$, such that $\rmCompDfa{R}{C} \bisim (\parallelComposition_{i=1}^N \rewardMachine{R}_i) \parallelComposition \dfa{C}$, and let $\marlEventSeq \in \marlEventSet^*$ be an $\mdp{M}$-attainable event sequence.

    By assumption, we have that $\marlEventSeq \models \varphi^\causalDiagram_\marlEventSet$.
    If $\rewardMachine{R}(\marlEventSeq) = 0$, then $\rmCompDfa{R}{C} = 0$, by definition of parallel composition.
    Due to $\rmCompDfa{R}{C} \bisim (\parallelComposition_{i=1}^N \rewardMachine{R}_i)\parallelComposition \dfa{C}$, we also have $(\parallelComposition_{i=1}^N \rewardMachine{R}_i)\parallelComposition \dfa{C} = 0$.
    Here, we used the fact that the pair of reward machines $\rmCompDfa{R}{C}$ and $(\parallelComposition_{i=1}^N \rewardMachine{R}_i)\parallelComposition \dfa{C}$ are bisimilar, and pass the \emph{strict} decomposition criterion~\ref{thm:decomposition}.
    Because $\dfa{C}(\xi) = 1$ (i.e., $\delta_{\dfa{C}}(\dfaInitState^C, \marlEventSeq) \in \dfaAccOf{C}$), and the definition of parallel composition, we have $(\parallelComposition_{i=1}^N \rewardMachine{R}_i)(\marlEventSeq) = 0$.
    Again by the definition of parallel composition, we have that $\exists i \in \mathset{1, \ldots, N} : \rewardMachine{R}_i(\marlProjection_i(\marlEventSeq)) = 0$.
    The converse direction proceeds analogously.
\end{proof}



\subsection{Proof of Theorem~\ref{thm:decomposition-decisions-agreement} (Criterion Compatibility)}
\label{subsec:proof-relaxed-criterion}

\begin{proof}
    The proof is direct. Assume $\bisimAlt \subseteq \rmStatesOf{R} \times \rmStatesOf{P}$ is a bisimulation between $\rewardMachine{R}$ and $\rewardMachine{P}$.

    Let $\dfaStatesOf{C}$ be the set of states of the causal DFA $\dfa{C}$.
    We define the relation $\tilde{\bisimAlt} \subseteq \rmStates_{\rmCompDfa{R}{C}} \times \rmStates_{\rmCompDfa{P}{C}}$ via Formula~\ref{eqn:tilde-bisim}, for all states $(\rmCommonStateOf{R}, \dfaCommonState_1) \in \rmStates_{\rmCompDfa{R}{C}}$ and $(\rmCommonStateOf{P}, \dfaCommonState_2) \in \rmStates_{\rmCompDfa{P}{C}}$.

    \begin{equation}
        ((\rmCommonStateOf{R}, \dfaCommonState_1), (\rmCommonStateOf{P}, \dfaCommonState_2)) \in \tilde{\bisimAlt} \Leftrightarrow (\rmCommonStateOf{R}, \rmCommonStateOf{P}) \in \bisimAlt \land \dfaCommonState_1 = \dfaCommonState_2
        \label{eqn:tilde-bisim}
    \end{equation}

    By checking the bisimulation conditions, we will show that $\tilde{\bisimAlt}$ is a bisimulation between $\rmCompDfa{R}{C}$ and $\rmCompDfa{P}{C}$.

    \begin{enumerate}
        \item $((\rmInitOf{R}, \dfaInitState), (\rmInitOf{P}, \dfaInitState)) \in \tilde{\bisimAlt}$ because $(\rmInitOf{R}, \rmInitOf{P}) \in \bisimAlt$.

        \item Let $((\rmCommonStateOf{R}, \dfaCommonState_1), (\rmCommonStateOf{P}, \dfaCommonState_2)) \in \tilde{\bisimAlt}$, and $e \in \marlEventSet$ some event.
              Then $(\rmCommonStateOf{R}, \rmCommonStateOf{P}) \in \bisimAlt$ and $\dfaCommonState_1 = \dfaCommonState_2 = \dfaCommonState \in \dfaStatesOf{C}$.
              Therefore, we have the following.
              \begin{itemize}
                  \item \textbf{(Accepting)}
                        $(\rmCommonStateOf{R}, \dfaCommonState) \in \dfaAcc^{\rmCompDfa{R}{C}}$ if and only if $(\rmCommonStateOf{P}, \dfaCommonState) \in \dfaAcc^{\rmCompDfa{P}{C}}$, because $(\rmCommonStateOf{R}, \dfaCommonState) \in \dfaAcc^{\rmCompDfa{R}{C}} \Leftrightarrow \rmCommonStateOf{R} \in \rmAccOf{R} \land \dfaCommonState \in \dfaAccOf{C} \Leftrightarrow \rmCommonStateOf{P} \in \rmAccOf{P} \land \dfaCommonState \in \dfaAccOf{C} \Leftrightarrow (\rmCommonStateOf{P}, \dfaCommonState) \in \dfaAcc^{\rmCompDfa{P}{C}}$.

                  \item \textbf{(Forth)}
                        if $\dfaTrans_{\rmCompDfa{R}{C}}((\rmCommonStateOf{R}, \dfaCommonState), e) = (\rmCommonStateOf{R}_{\star}, \dfaCommonState_{\star})$, then there exists $\rmCommonStateOf{P}_{\star}$ such that $\dfaTrans_{\rmCompDfa{P}{C}}((\rmCommonStateOf{P}, \dfaCommonState), e) = (\rmCommonStateOf{P}_{\star}, \dfaCommonState_{\star})$ and \\$((\rmCommonStateOf{R}_{\star}, \dfaCommonState_{\star}), (\rmCommonStateOf{P}_{\star}, \dfaCommonState_{\star})) \in \tilde{\bisimAlt}$.

                  \item \textbf{(Back)}
                        if $\dfaTrans_{\rmCompDfa{P}{C}}((\rmCommonStateOf{P}, \dfaCommonState), e) = (\rmCommonStateOf{P}_{\star}, \dfaCommonState_{\star})$, then there exists $\rmCommonStateOf{R}_{\star}$ such that $\dfaTrans_{\rmCompDfa{R}{C}}((\rmCommonStateOf{R}, \dfaCommonState), e) = (\rmCommonStateOf{R}_{\star}, \dfaCommonState_{\star})$ and \\$((\rmCommonStateOf{R}_{\star}, \dfaCommonState_{\star}), (\rmCommonStateOf{P}_{\star}, \dfaCommonState_{\star})) \in \tilde{\bisimAlt}$.
              \end{itemize}
    \end{enumerate}

    One obtains $\rmCommonState^{\rewardMachine{P}}_{\star}$ ($\rmCommonState^{\rewardMachine{R}}_{\star}$) from the bisimulation condition on $\rewardMachine{P}$ and $\rewardMachine{R}$.
    Then, $((\rmCommonState^{\rewardMachine{P}}_{\star}, \dfaCommonState_{\star}), (\rmCommonState^{\rewardMachine{R}}_{\star}, \dfaCommonState_{\star})) \in \tilde{\bisimAlt}$ by definition of $\tilde{\bisimAlt}$.
    The diagrams in Figure~\ref{fig:proof-diagram-a}, Figure~\ref{fig:proof-diagram-b}, and Figure~\ref{fig:proof-diagram-c} illustrate the back and forth bisimulation conditions.

\begin{figure}[ht]
    \centering
    \begin{subfigure}{.45\columnwidth}
        \centering
        \resizebox{\columnwidth}{!}{\begin{tikzpicture}[>=Stealth, node distance=2.5cm and 2.5cm]
    \tikzstyle{state}=[circle, fill=black, inner sep=0pt, minimum size=5pt]

    \node[state, label=above:{$\rmCommonStateOf{R}$}] (A) {};
    \node[state, label=above:{$\rmCommonStateOf{P}$}] (B) [right=of A] {};
    \node[state, label=below:{$\rmCommonStateOf{R}_{\star}$}] (C) [below=of A] {};
    \node[state, label=below:{$\rmCommonStateOf{P}_{\star}$}] (D) [below=of B] {};

    \draw[->] (A) -- (B) node[midway, above] {$\bisimAlt$};
    \draw[->, blue, bend right=30] (A) to node[midway, left] {$\dfaTrans_{\rewardMachine{R}}(e)$} (C);
    \draw[->, blue, bend right=30, dashed] (B) to node[midway, left] {$\dfaTrans_{\rewardMachine{P}}(e)$} (D);
    \draw[->, dashed] (C) -- (D) node[midway, below] {$\bisimAlt$};
\end{tikzpicture}}
        \caption{Forth}
        \label{fig:proof-diagram-forth-a}
    \end{subfigure}%
    \hfill
    \begin{subfigure}{.45\columnwidth}
        \centering
        \resizebox{\columnwidth}{!}{\begin{tikzpicture}[>=Stealth, node distance=2.5cm and 2.5cm]
    \tikzstyle{state}=[circle, fill=black, inner sep=0pt, minimum size=5pt]

    \node[state, label=above:{$\rmCommonStateOf{R}$}] (A) {};
    \node[state, label=above:{$\rmCommonStateOf{P}$}] (B) [right=of A] {};
    \node[state, label=below:{$\rmCommonStateOf{R}_{\star}$}] (C) [below=of A] {};
    \node[state, label=below:{$\rmCommonStateOf{P}_{\star}$}] (D) [below=of B] {};

    \draw[->] (A) -- (B) node[midway, above] {$\bisimAlt$};
    \draw[->, blue, bend right=30, dashed] (A) to node[midway, left] {$\dfaTrans_{\rewardMachine{R}}(e)$} (C);
    \draw[->, blue, bend right=30] (B) to node[midway, left] {$\dfaTrans_{\rewardMachine{P}(e)}$} (D);
    \draw[->, dashed] (C) -- (D) node[midway, below] {$\bisimAlt$};
\end{tikzpicture}}
        \caption{Back}
        \label{fig:proof-diagram-back-a}
    \end{subfigure}
    \caption{Visualizing $\rewardMachine{R} \bisim \rewardMachine{P}$.}
    \label{fig:proof-diagram-a}
\end{figure}

\begin{figure}[ht]
    \centering
    \begin{subfigure}{.45\columnwidth}
        \centering
        \resizebox{\columnwidth}{!}{\begin{tikzpicture}[>=Stealth, node distance=2.5cm and 2.5cm]
    \tikzstyle{state}=[circle, fill=black, inner sep=0pt, minimum size=5pt]

    \node[state, label=above:{$\rmCommonStateOf{R}, \dfaCommonState$}] (A) {};
    \node[state, label=above:{$\rmCommonStateOf{P}, \dfaCommonState$}] (B) [right=of A] {};
    \node[state, label=below:{$\rmCommonStateOf{R}_{\star}, \dfaCommonState_{\star}$}] (C) [below=of A] {};
    \node[state, label=below:{$\rmCommonStateOf{P}_{\star}, \dfaCommonState_{\star}$}] (D) [below=of B] {};

    \draw[->] (A) -- (B) node[midway, above] {$\bisimAlt$, $=$};
    \draw[->, blue, bend right=30] (A) to node[near start, left] {$\dfaTrans_{\rewardMachine{R}}(\marlAtomicEvent)$} (C);
    \draw[->, red, bend left=30] (A) to node[near end, right] {$\dfaTrans_{\dfa{C}}(\marlAtomicEvent)$} (C);
    \draw[->, blue, bend right=30, dashed] (B) to node[near start, left] {$\dfaTrans_{\rewardMachine{P}}(\marlAtomicEvent)$} (D);
    \draw[->, red, bend left=30, dashed] (B) to node[near end, right] {$\dfaTrans_{\dfa{C}}(\marlAtomicEvent)$} (D);
    \draw[->, dashed] (C) -- (D) node[midway, below] {$\bisimAlt$, $=$};
\end{tikzpicture}}
        \caption{Forth}
        \label{fig:proof-diagram-forth-b}
    \end{subfigure}%
    \hfill
    \begin{subfigure}{.45\columnwidth}
        \centering
        \resizebox{\columnwidth}{!}{\begin{tikzpicture}[>=Stealth, node distance=2.5cm and 2.5cm]
    \tikzstyle{state}=[circle, fill=black, inner sep=0pt, minimum size=5pt]

    \node[state, label=above:{$\rmCommonStateOf{R}, \dfaCommonState$}] (A) {};
    \node[state, label=above:{$\rmCommonStateOf{P}, \dfaCommonState$}] (B) [right=of A] {};
    \node[state, label=below:{$\rmCommonStateOf{R}_{\star}, \dfaCommonState_{\star}$}] (C) [below=of A] {};
    \node[state, label=below:{$\rmCommonStateOf{P}_{\star}, \dfaCommonState_{\star}$}] (D) [below=of B] {};

    \draw[->] (A) -- (B) node[midway, above] {$\bisimAlt$, $=$};
    \draw[->, blue, bend right=30, dashed] (A) to node[near start, left] {$\dfaTrans_{\rewardMachine{R}}(e)$} (C);
    \draw[->, red, bend left=30, dashed] (A) to node[near end, right] {$\dfaTrans_{\dfa{C}}(e)$} (C);
    \draw[->, blue, bend right=30] (B) to node[near start, left] {$\dfaTrans_{\rewardMachine{P}}(e)$} (D);
    \draw[->, red, bend left=30] (B) to node[near end, right] {$\dfaTrans_{\dfa{C}}(e)$} (D);
    \draw[->, dashed] (C) -- (D) node[midway, below] {$\bisimAlt$, $=$};
\end{tikzpicture}}
        \caption{Back}
        \label{fig:proof-diagram-back-b}
    \end{subfigure}
    \caption{Visualizing $\rewardMachine{R} \bisim \rewardMachine{P}$ alongside transitions from $\dfa{C}$.}
    \label{fig:proof-diagram-b}
\end{figure}
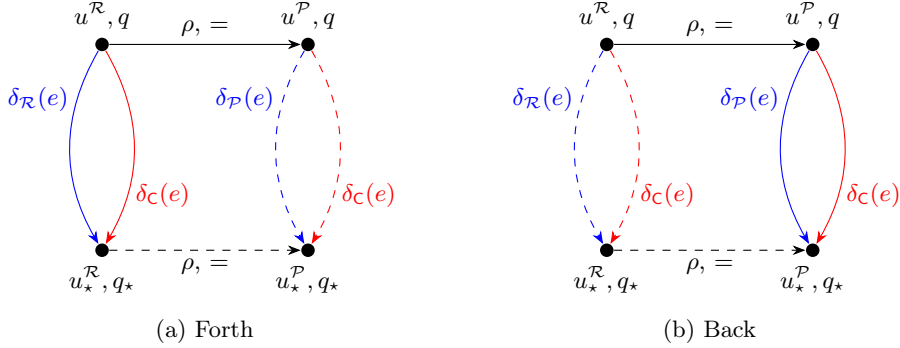

\begin{figure}[ht]
    \centering
    \begin{subfigure}{.45\columnwidth}
        \centering
        \resizebox{\columnwidth}{!}{\begin{tikzpicture}[>=Stealth, node distance=2.5cm and 2.5cm]
    \tikzstyle{state}=[circle, fill=black, inner sep=0pt, minimum size=5pt]

    \node[state, label=above:{$(\rmCommonStateOf{R}, \dfaCommonState)$}] (A) {};
    \node[state, label=above:{$(\rmCommonStateOf{P}, \dfaCommonState)$}] (B) [right=of A] {};
    \node[state, label=below:{$(\rmCommonStateOf{R}_{\star}, \dfaCommonState_{\star})$}] (C) [below=of A] {};
    \node[state, label=below:{$(\rmCommonStateOf{P}_{\star}, \dfaCommonState_{\star})$}] (D) [below=of B] {};

    \draw[->] (A) -- (B) node[midway, above] {$\bisimAlt$};
    \draw[->] (A) to node[midway, left] {$\dfaTrans_{\rmCompDfa{R}{C}}(e)$} (C);
    \draw[->, dashed] (B) to node[midway, left] {$\dfaTrans_{\rmCompDfa{P}{C}}(e)$} (D);
    \draw[->, dashed] (C) -- (D) node[midway, below] {$\bisimAlt$};
\end{tikzpicture}}
        \caption{Forth}
        \label{fig:proof-diagram-forth-c}
    \end{subfigure}%
    \hfill
    \begin{subfigure}{.45\columnwidth}
        \centering
        \resizebox{\columnwidth}{!}{\begin{tikzpicture}[>=Stealth, node distance=2.5cm and 2.5cm]
    \tikzstyle{state}=[circle, fill=black, inner sep=0pt, minimum size=5pt]

    \node[state, label=above:{$(\rmCommonStateOf{R}, \dfaCommonState)$}] (A) {};
    \node[state, label=above:{$(\rmCommonStateOf{P}, \dfaCommonState)$}] (B) [right=of A] {};
    \node[state, label=below:{$(\rmCommonStateOf{R}_{\star}, \dfaCommonState_{\star})$}] (C) [below=of A] {};
    \node[state, label=below:{$(\rmCommonStateOf{P}_{\star}, \dfaCommonState_{\star})$}] (D) [below=of B] {};

    \draw[->] (A) -- (B) node[midway, above] {$\bisimAlt$};
    \draw[->, dashed] (A) to node[midway, left] {$\dfaTrans_{\rmCompDfa{R}{C}}(e)$} (C);
    \draw[->] (B) to node[midway, left] {$\dfaTrans_{\rmCompDfa{P}{C}}(e)$} (D);
    \draw[->, dashed] (C) -- (D) node[midway, below] {$\bisimAlt$};
\end{tikzpicture}}
        \caption{Back}
        \label{fig:proof-diagram-back-c}
    \end{subfigure}
    \caption{Visualizing $\rmCompDfa{R}{C} \bisim \rmCompDfa{P}{C}$.}
    \label{fig:proof-diagram-c}
\end{figure}

\end{proof}

\subsection{Proof of Theorem~\ref{thm:decomposition-viability} (Relaxed Decomposition Viability)}
\label{subsec:proof-decomposition-viability}

\begin{proof}
    Let $\mdp{M} = \langle \mdpStates, \mdpInit, \mdpActions, \mdpTrans, \mdpDiscount, \rewardMachine{R}, \mdpLabel \rangle$ be the team task RM-MDP, with local event sets $\marlEventSet_1, \ldots, \marlEventSet_N$ and decomposable label function $\mdpLabel$, $\mdpLabel_1, \ldots, \mdpLabel_N$ the local labeling functions of agents $1, \ldots, N$, and $\causalDiagram$ the TL-CD which holds for $\mdp{M}$.
    Assume that the relaxed decomposition criterion holds, and that agents synchronize on shared events.

    We use $\langle \mdpLabel_i(\mdpCommonState_0^i \rmCommonState_0^i \cdots \mdpCommonState_k^i \rmCommonState_k^i) \mid \text{Sync}^i \rangle$ to denote the synchronized local labeling function output.
    $\text{Sync}^i$ is a binary vector, and $\text{Sync}^i_t = 1$ iff. agent $i$ received a synchronization signal in time-step $t$.

    For the first claim we need to show that for all team trajectories $\mdpCommonState_0 \rmCommonState_0 \cdots \mdpCommonState_k \rmCommonState_k$ and local trajectories $\mathset{\mdpCommonState_0^i \rmCommonState_0^i \cdots \mdpCommonState_k^i \rmCommonState_k^i}_{i=1}^N$, we have $\rewardMachine{R}(\mdpLabel(\mdpCommonState_0 \rmCommonState_0 \cdots \mdpCommonState_k \rmCommonState_k)) = 1$ if and only if $\rewardMachine{R}_i(\langle \mdpLabel_i(\mdpCommonState_0^i \rmCommonState_0^i \cdots \mdpCommonState_k^i \rmCommonState_k^i) \mid \text{Sync}^i \rangle) = 1$ for all $i = 1, \ldots, N$.
    
    We first note that $\mdpCommonState_0 \rmCommonState_0 \cdots \mdpCommonState_k \rmCommonState_k$ is sampled from $\mdp{M}$, which means that $\mdpLabel(\mdpCommonState_0 \rmCommonState_0 \cdots \mdpCommonState_k \rmCommonState_k) \models \varphi^\causalDiagram_\marlEventSet$.
    Then the result follows from Theorem~\ref{thm:decomposition-relaxed} if we show that $\marlProjection_i(\mdpLabel(\teamTrajectory)) = \langle \mdpLabel_i(\localStatesTrajectorySingle) \mid \text{Sync}^i \rangle$ (for every $i = 1, \ldots, N$).
    Let $\mdpCommonLabel_t$ denote the output of $\mdpLabel$ at time-step $t = 0, \ldots, k$, $\mdpCommonLabel_t^i$ the output of $\mdpLabel_i$, and $\tilde{\mdpCommonLabel_t^i}$ the syncrhonized output of $\mdpLabel_i$.

    We proceed by induction over time-steps $t$.

    \textbf{Base case:} $t = 0$.
    In that case, by definition, $\mdpCommonLabel_0 = \mdpCommonLabel^i_0 = \tilde{\mdpCommonLabel^i_0} = \emptyset$.
    Additionally, by definition, $\rmCommonState_0 = \rmInit \in \rmInit^i = \rmCommonState^i_0$ for all $i = 1, \ldots, N$ (the initial state $\rmCommonState_0$ of $\rewardMachine{R}$ is matched by the initial states $\rmCommonState^i_0$ of all projections $\rewardMachine{R}_i$).

    \textbf{Assumption:}
    Let $t \in \naturals$, $0 < t < k$ such that for all $i = 1, \ldots, N$ it holds that $\mdpCommonLabel_{t} \cap \marlEventSet_i = \tilde{\mdpCommonLabel^i_{t}}$,
    and $\rmCommonState_{t} \in \rmCommonState^i_{t}$.

    \textbf{Inductive step:}
    By the inductive hypothesis $\rmCommonState_{t} \in \rmCommonState^i_{t}$.
    By the decomposability of $\mdpLabel$, the output of $\mdpLabel_i(\mdpCommonState^i_{t}, \rmCommonState^i_{t}, \mdpCommonState^i_{t+1}) = \mdpCommonLabel^i_{t+1}$ is well-defined.
    There are three cases to consider: (1) $\mdpCommonLabel^i_{t+1} = \mdpCommonLabel_{t+1} \cap \marlEventSet_i = \mathset{\marlAtomicEvent}$ and $|\marlShared_{\marlAtomicEvent}| > 1$; (2) $\mdpCommonLabel^i_{t+1} = \mdpCommonLabel_{t+1} \cap \marlEventSet_i = \mathset{\marlAtomicEvent}$ and $|\marlShared_{\marlAtomicEvent}| = 1$; and (3) $\mdpCommonLabel^i_{t+1} = \emptyset$.
    
    In cases (2) and (3), we immediately have $\mdpCommonLabel^i_{t+1} = \mdpCommonLabel_{t+1} \cap \marlEventSet_i = \tilde{\mdpCommonLabel^i_{t+1}}$ (no synchronization necessary), and by definition of parallel projections and the inductive hypothesis, $\rmTrans(\rmCommonState_{t}, \mdpCommonLabel_{t+1}) = \rmCommonState_{t+1} \in \rmCommonState^i_{t+1} = \rmTrans^i(\rmCommonState^i_{t}, \tilde{\mdpCommonLabel^i_{t+1}})$.
    
    In case (1), for all $i \in \marlShared_\marlAtomicEvent$ we have $\text{Sync}^i_{t+1} = 1$ if $\marlAtomicEvent \in \mdpCommonLabel_{t+1}$, or $\text{Sync}^i_{t+1} = 0$ otherwise.
    If $\text{Sync}^i_{t+1} = 1$, then $\tilde{\mdpCommonLabel^i_{t+1}} = \mathset{e}$, otherwise $\tilde{\mdpCommonLabel^i_{t+1}} = \emptyset$.
    The rest of case (1) follows analogously to cases (2) and (3), and we may conclude that $\rmCommonState_{t+1} \in \rmTrans^i(\rmCommonState^i_{t}, \tilde{\mdpCommonLabel^i_{t+1}})$ for all $i = 1, \ldots, N$ in all cases, i.e. the $\marlIndistinguishability_i$-classes of the projected RMs match the state of the team RM in all steps.

    That, in turn, implies $\marlProjection_i(\mdpLabel(\teamTrajectory)) = \langle \mdpLabel_i(\localStatesTrajectorySingle) \mid \text{Sync}^i \rangle$ for all $i = 1, \ldots, N$, and the first claim is proven via Theorem~\ref{thm:decomposition-relaxed} and $\mdpLabel(\teamTrajectory) \models \varphi^\causalDiagram_\marlEventSet$.
    
    The second claim is that $\max\mathset{0, \valueFunction^{\policy}_1(\mdpInit) + \cdots + \valueFunction^{\policy}_N(\mdpInit) - (N - 1)} \leq \valueFunction^{\policy}(\mdpInit) \leq \min\mathset{\valueFunction^{\policy}_1(\mdpInit), \ldots, \valueFunction^{\policy}(\mdpInit)}$, where $\valueFunction^{\policy}(\mdpInit)$ denotes the team success probability, and $\valueFunction^{\policy}_i(\mdpInit)$ the success probability of agent $i$.
    We use the notation $\valueFunction^{\policy}$ to evoke the correspondence between the success probability and the value function in non-discounted task-completion RM-MDPs ($\mdpGamma = 1$).

    By the previous result, for any sequence of team states $\teamTrajectory$, we have $\rewardMachine{R}(\mdpLabel(\teamTrajectory)) = 1$ if and only if $\rewardMachine{R}_i(\langle \mdpLabel_i(\localStatesTrajectorySingle) \mid \text{Sync}^i \rangle) = 1$ for all $i = 1, \ldots, N$.
    This implies that the probability of completing the task captured by $\rewardMachine{R}$ under policy $\policy$ is equal to the probability of simultaneously completing all tasks captured by $\rewardMachine{R}_i$ (when $\policy = (\policy_1, \ldots, \policy_N)$).
    Then the second claim follows by applying the Fr\'echet conjunction inequality to the success probabilities of the agents.
\end{proof}

\section{Case Study 3: Buttons Task}
\label{sec:case_study_2}

\begin{figure}[ht]
    \centering
    \begin{subfigure}{.45\textwidth}
        \centering
        \resizebox{0.7\linewidth}{!}{\scalebox{0.3}{ 
    \begin{tikzpicture}[scale=0.6] 
    \draw (0,0) grid (12,12); 

    \fill[black!70] (2,1) rectangle (3,12); 
    \fill[black!70] (2,1) rectangle (12,2); 
    \fill[black!70] (6,2) rectangle (7,4); 
    \fill[black!70] (6,3) rectangle (8,4); 
    \fill[black!70] (8,3) rectangle (9,8); 
    \fill[black!70] (10,7) rectangle (11,12); 

    \fill[yellow!50] (1,11) rectangle (2,12); 
\fill[green!50] (3,5) rectangle (4,6); 
\fill[green!50] (7,2) rectangle (8,3); 
\fill[red!50] (11,2) rectangle (12,3); 

\fill[yellow!50] (3,9) rectangle (10,10); 
\fill[green!50] (12,10) rectangle (11,8);
\fill[red!50] (7,0) rectangle (9,1);

    \node at (0.5,6.5) {\texttt{A1}};
    \node at (3.5,11.5) {\texttt{A2}};
    \node at (11.5,11.5) {\texttt{A3}};
    \node at (1.5,11.5) {\texttt{B}};
    \node at (3.5,5.5) {\texttt{B1}};
    \node at (7.5,2.5) {\texttt{B2}};
    \node at (11.5,2.5) {\texttt{B3}};
    \node at (11.5,0.5) {\gridGoal{}};

    \node at (5.5, 5.5) {\tikz[baseline=(char.base)]{\node[line width=1pt,blue,draw,circle,inner sep=0.1pt, text=black] (char) {S};}};
    \node at (7.5, 5.5) {\tikz[baseline=(char.base)]{\node[line width=1pt,blue,draw,circle,inner sep=0.1pt, text=black] (char) {S};}};

\draw[->, blue, thick, line width=1.5pt] (3.5,7.8) -- (3.5,7.3);  
\draw[->, blue, thick, line width=1.5pt] (4.5,7.8) -- (4.5,7.3);  
\draw[->, blue, thick, line width=1.5pt] (5.5,7.8) -- (5.5,7.3);  
\draw[->, blue, thick, line width=1.5pt] (6.5,7.8) -- (6.5,7.3); 
\draw[->, blue, thick, line width=1.5pt] (7.5,7.8) -- (7.5,7.3); 
\draw[->, blue, thick, line width=1.5pt] (9.5,7.8) -- (9.5,7.3);  
\end{tikzpicture}

}}
        \caption{$B$, $B_1$, $B_2$, and $B_3$ are buttons, and \gridGoal{} is the goal. One-way doors are represented by the downwards arrow.
            States labeled with $S$ are signal lights which indicate that the agent is in the region that cannot go to the red button.}
        \label{fig:case-study-2-gridworld}
    \end{subfigure}%
    \hfill
    \begin{subfigure}{.45\textwidth}
        \centering
        \resizebox{1\linewidth}{!}{\begin{tikzpicture}[>=stealth, node distance=1.2cm, every state/.append style={scale=1.0}, initial text=]
    \node[state, initial] (uI) at (0, 0) {$u_I$};
    \node[state] (u1) at (2, 0) {$u_1$};
    \node[state] (u2) at (4, 0) {$u_2$};
    \node[state] (u3) at (6, 1.5) {$u_3$};
    \node[state] (u4) at (6, -1.5) {$u_4$};
    \node[state] (u5) at (8, 0) {$u_5$};
    \node[state] (u6) at (10, 0) {$u_6$};
    \node[state, accepting] (u7) at (10, -1.5) {$u_7$};

    \path[->] (uI) edge node[above, font=\small] {$B$} (u1);
    \path[->] (u1) edge node[above, font=\small] {$B_2 \lor B_1$} (u2);
    \path[->] (u2) edge[bend left] node[above, pos=0.5, font=\small, anchor=south] {$A_2^{\neg B_3}$} (u3);
    \path[->] (u2) edge[bend right] node[below, pos=0.5, font=\small, anchor=north] {$A_3^{\neg B_3}$} (u4);
    \path[->] (u3) edge[bend left] node[above, pos=0.5, font=\small, anchor=south] {$A_3^{B_3}$} (u5);
    \path[->] (u4) edge[bend right] node[below, pos=0.5, font=\small, anchor=north] {$A_2^{B_3}$} (u5);
    \path[->] (u5) edge node[above, font=\small] {$B_3$} (u6);
    \path[->] (u6) edge node[right, font=\small] {Goal} (u7);
    \path[->] (u3) edge node[below, pos=0.5, font=\small] {$A_2^{B_3}$} (u2);
    \path[->] (u4) edge node[above, pos=0.5, font=\small] {$A_3^{B_3}$} (u2);
    \path[->] (u5) edge node[below, pos=0.5, font=\small] {$A_2^{\neg B_3}$} (u3);
    \path[->] (u5) edge node[above, pos=0.5, font=\small] {$A_3^{\neg B_3}$} (u4);
    \path[->] (-1.5, 0) edge[->] node[above, font=\small] {start} (uI);
\end{tikzpicture}}
        \caption{Reward machine encoding the cooperative buttons task. Output is $1$ on transitions into accepting states, and $0$ otherwise.}
        \label{fig:case-study-2-team-rm}
    \end{subfigure}
    \caption{Environment (left) and RM (right) for the \emph{cooperative buttons} task.
        One-way doors block agent $1$ to go to the red button and push the button.}
    \label{fig:case-study-2}
\end{figure}
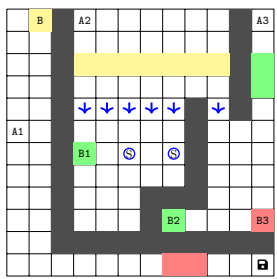
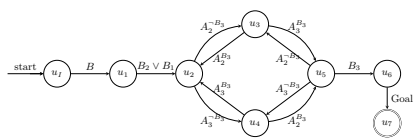

In our third case study, we perform the same analysis for a 3-agent problem we call the \emph{Cooperative Buttons} Task.
In this environment shown in Figure~\ref{fig:case-study-2-gridworld}, events $B$ , $B_1$ , $B_2$ , and $B_3$ correspond to yellow, green, and red buttons respectively.
In this shared environment, there are three agents - A1, A2, and A3 - who have to work together to
achieve the objective of allowing A1 to reach the goal location known as Goal. However, the path to
the goal is blocked by three colored regions - red, yellow, and green - which correspond to the paths
of agents A1, A2, and A3 respectively. To cross these colored regions, the corresponding colored
button must be pressed first. The yellow and green buttons can be pressed by an individual agent,
but the red button requires two agents to simultaneously occupy the button's location before it gets
activated.

In order to complete the task, the agents have to follow a specific sequence of events. First, A1 should
push the yellow button, which will allow A2 to proceed to the green button. Pressing the green button
is necessary for A3 to join A2 in pressing the red button, which will finally allow A1 to cross the red
region and reach the goal location. The sequence of events is encoded in the reward machine depicted in Figure~\ref{fig:case-study-2-team-rm}.

\begin{figure}[h]
  \centering
  \begin{tikzpicture}
    \begin{axis}[
        width=0.49\linewidth, 
        grid=major,
        grid style={dashed,gray!30},
        xlabel=Training Steps,
        ylabel=Steps to Task Completion,
        legend style={
            at={(0.03,0.03)},
            anchor=south west,
            legend cell align=left,
            draw=black,
            fill=white,
            text opacity=1,
            opacity=0.8,
            font=\scriptsize, 
            inner sep=2pt, 
            outer sep=1pt, 
          },
        x tick label style={rotate=45,anchor=east},
        tick label style={font=\small}, 
        label style={font=\small}, 
        ytick={0,200,400,600,800,1000},  
        title={\textbf{Decentralized (Buttons)}},
      ]

      \addplot[
        thick,
        red
      ] table [y=prc_50, x=steps, col sep=comma] {results/final/buttons_no_tlcd/2024-05-22_15-44-15.csv};
      \addlegendentry{No TL-CD}

      \addplot[
        name path=upper,
        draw=none
      ] table[y=prc_75, x=steps, col sep=comma] {results/final/buttons_no_tlcd/2024-05-22_15-44-15.csv};

      \addplot[
        name path=lower,
        draw=none
      ] table[y=prc_25, x=steps, col sep=comma] {results/final/buttons_no_tlcd/2024-05-22_15-44-15.csv};

      \addplot [
        fill=red,
        fill opacity=0.25
      ] fill between[of=upper and lower];

      \addplot[
        thick,
        blue
      ] table [y=prc_50, x=steps, col sep=comma] {results/final/buttons_tlcd/server_results.csv};
      \addlegendentry{TL-CD}

      \addplot[
        name path=upper_tlcd,
        draw=none
      ] table[y=prc_75, x=steps, col sep=comma] {results/final/buttons_tlcd/server_results.csv};

      \addplot[
        name path=lower_tlcd,
        draw=none
      ] table[y=prc_25, x=steps, col sep=comma] {results/final/buttons_tlcd/server_results.csv};

      \addplot [
        fill=blue,
        fill opacity=0.25
      ] fill between[of=upper_tlcd and lower_tlcd];

    \end{axis}
  \end{tikzpicture}
  \hspace{3mm}%
  \begin{tikzpicture}
    \begin{axis}[
        width=0.49\linewidth, 
        grid=major,
        grid style={dashed,gray!30},
        xlabel=Training Steps,
        legend style={
            at={(0.03,0.03)},
            anchor=south west,
            legend cell align=left,
            draw=black,
            fill=white,
            text opacity=1,
            opacity=0.8,
            font=\scriptsize, 
            inner sep=2pt, 
            outer sep=1pt, 
          },
        x tick label style={rotate=45,anchor=east},
        tick label style={font=\small},
        label style={font=\small},
        ytick={0,200,400,600,800,1000},  
        title={\textbf{Centralized (Buttons)}},
      ]

      \addplot[
        thick,
        red
      ] table [y=prc_50, x=steps, col sep=comma] {results/final/centralized_buttons_no_tlcd/2024-05-22_11-52-05.csv};
      \addlegendentry{No TL-CD}

      \addplot[
        name path=upper,
        draw=none
      ] table[y=prc_75, x=steps, col sep=comma] {results/final/centralized_buttons_no_tlcd/2024-05-22_11-52-05.csv};

      \addplot[
        name path=lower,
        draw=none
      ] table[y=prc_25, x=steps, col sep=comma] {results/final/centralized_buttons_no_tlcd/2024-05-22_11-52-05.csv};

      \addplot [
        fill=red,
        fill opacity=0.25
      ] fill between[of=upper and lower];

      \addplot[
        thick,
        blue
      ] table [y=prc_50, x=steps, col sep=comma] {results/final/centralized_buttons_tlcd/2024-05-22_11-13-56.csv};
      \addlegendentry{TL-CD}

      \addplot[
        name path=upper_tlcd,
        draw=none
      ] table[y=prc_75, x=steps, col sep=comma] {results/final/centralized_buttons_tlcd/2024-05-22_11-13-56.csv};

      \addplot[
        name path=lower_tlcd,
        draw=none
      ] table[y=prc_25, x=steps, col sep=comma] {results/final/centralized_buttons_tlcd/2024-05-22_11-13-56.csv};

      \addplot [
        fill=blue,
        fill opacity=0.25
      ] fill between[of=upper_tlcd and lower_tlcd];

    \end{axis}
  \end{tikzpicture}

  \caption{Buttons Task Comparisons. Results of $10$ independent runs ($2$ in the case of the Centralized algorithm, which does not converge in $10^6$ steps).}
  \label{fig:case-study-2-results}
\end{figure}
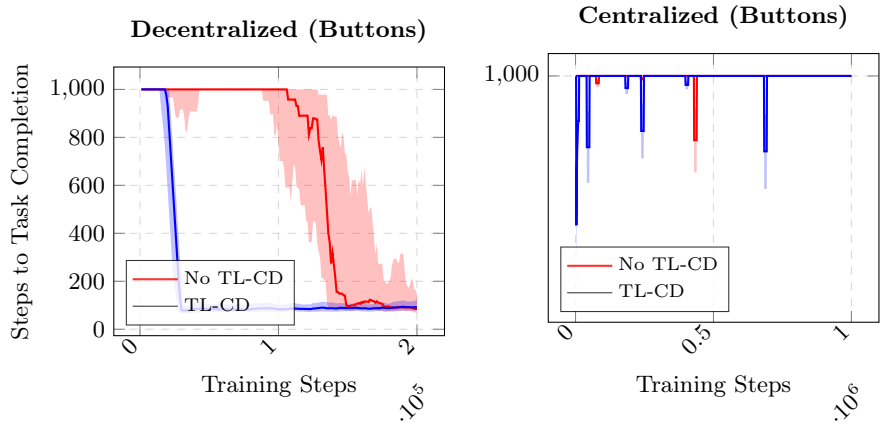

The results for this case study are shown in Figure~\ref{fig:case-study-2-results}.

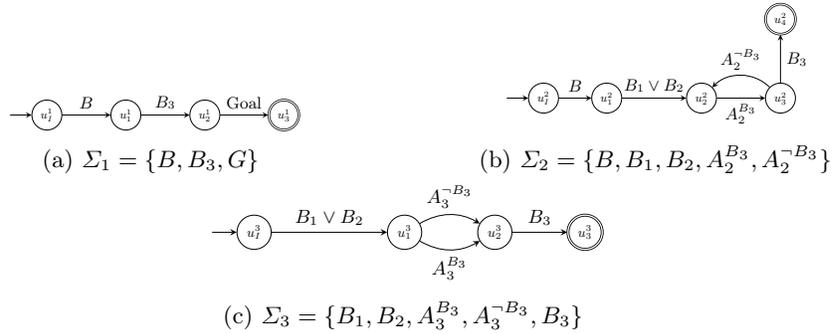
\begin{figure}[ht]
    \centering
    \begin{subfigure}{.45\columnwidth}
        \centering
        \scalebox{0.7}{\begin{tikzpicture}[>=stealth, node distance=0.5cm, every state/.append style={scale=0.7}, initial text=]
    \node[state, initial] (0) at (0, 0) {$u_I^1$};
    \node[state] (1) at (1.5, 0) {$u_1^1$};
    \node[state] (2) at (3, 0) {$u_2^1$};
    \node[state, accepting] (3) at (4.5, 0) {$u_3^1$};

    \path[->] (0) edge node[above, font=\small] {$B$} (1);
    \path[->] (1) edge node[above, font=\small] {$B_3$} (2);
    \path[->] (2) edge node[above, font=\small] {Goal} (3);
\end{tikzpicture}}
        \caption{$\marlEventSet_1 = \mathset{B, B_3, G}$}
        \label{fig:case-study-2-projection-1}
    \end{subfigure}%
    \hfill
    \begin{subfigure}{.45\columnwidth}
        \centering
        \scalebox{0.7}{\begin{tikzpicture}[>=stealth, node distance=1cm, every state/.append style={scale=0.7}, initial text=]
    \node[state, initial] (0) at (0, 0) {$u_I^2$};
    \node[state] (1) at (1.2, 0) {$u_1^2$};
    \node[state] (2) at (3, 0) {$u_2^2$};
    \node[state] (3) at (4.5, 0) {$u_3^2$};
    \node[state, accepting] (4) at (4.5, 1.5) {$u_4^2$};

    \path[->] (0) edge node[above, font=\small] {$B$} (1);
    \path[->] (1) edge node[above, font=\small] {$B_1 \lor B_2$} (2);
    \path[->] (2) edge node[below, font=\small] {$A_2^{B_3}$} (3);
    \path[->] (3) edge node[right, font=\small] {$B_3$} (4);
    \path[->] (3) edge[bend right=45] node[above, font=\small] {$A_2^{\neg B_3}$} (2);
\end{tikzpicture}}
        \caption{$\marlEventSet_2 = \mathset{B, B_1, B_2, A_2^{B_3}, A_2^{\neg B_3}}$}
        \label{fig:case-study-2-projection-2}
    \end{subfigure}
    \vfill
    \begin{subfigure}{.9\columnwidth}
        \centering
        \scalebox{0.8}{\begin{tikzpicture}[>=stealth, node distance=1.5cm, every state/.append style={scale=0.7}, initial text=]
    \node[state, initial] (0) at (-1, 0) {$u_I^3$};
    \node[state] (1) at (1.5, 0) {$u_1^3$};
    \node[state] (2) at (3, 0) {$u_2^3$};
    \node[state, accepting] (3) at (4.5, 0) {$u_3^3$};

    \path[->] (0) edge node[above, font=\small] {$B_1 \lor B_2$} (1);
    \path[->] (1) edge[bend left] node[above, font=\small] {$A_3^{\neg B_3}$} (2);
    \path[->] (1) edge[bend right] node[below, font=\small] {$A_3^{B_3}$} (2);
    \path[->] (2) edge node[above, font=\small] {$B_3$} (3);
\end{tikzpicture}}
        \caption{$\marlEventSet_3 = \mathset{B_1, B_2, A_3^{B_3}, A_3^{\neg B_3}, B_3}$}
        \label{fig:case-study-2-projection-3}
    \end{subfigure}
    \caption{Projections of the team task RM from Figure~\ref{fig:case-study-2-team-rm} along local event sets of agents 1, 2, and 3.}
    \label{fig:case-study-2-projections}
\end{figure}

Figures ~\ref{fig:case-study-2-projection-1}, ~\ref{fig:case-study-2-projection-2}, and ~\ref{fig:case-study-2-projection-3}
show the results of projecting the task RM from Figure~\ref{fig:case-study-2-team-rm} onto the local event sets of $\marlEventSet_1$, $\marlEventSet_2$, and $\marlEventSet_3$, respectively.

\begin{figure}[ht]
    \centering
    \begin{subfigure}[t]{0.45\columnwidth}
        \centering
        \scalebox{0.7}{\begin{tikzpicture}[>=stealth, node distance=1.5cm, every node/.style={thick, draw, ellipse, align=center, inner sep=1pt}]
    \node[minimum width=1cm, minimum height=1cm] (B1) {$S$};
    \node[minimum width=1cm, minimum height=1cm, right=of B1] (B2) {$\globallyOp \lnot G$};

    \draw[->, thick] (B1) -- (B2);
\end{tikzpicture}}
        \caption{$\varphi^\causalDiagram \equiv \globallyOp(S \rightarrow \globallyOp \lnot G)$}
        \label{fig:case-study-2-tlcd-2}
    \end{subfigure}%
    \hfill
    \begin{subfigure}[t]{0.45\columnwidth}
        \centering
        \scalebox{0.7}{\begin{tikzpicture}[>=stealth, node distance=1.5cm, every state/.append style={scale=0.8}, initial text=]

    \node[state, initial, accepting] (0) at (0, 0) {$\dfaInitState$};
    \node[state, accepting] (1) at (2, 0) {$\dfaCommonState_1$};
    \node[state] (2) at (4, 0) {$\dfaCommonState_2$};

    \draw[->] (0) -- (1) node[midway, above] {$S$};
    \draw[->] (1) -- (2) node[midway, above] {$G$};

    \draw[->] (0) edge [loop above] node {$\lnot S$} ();
    \draw[->] (1) edge [loop above] node {$\lnot G$} ();
    \draw[->] (2) edge [loop above] node {$\top$} ();
\end{tikzpicture}}
        \caption{Causal DFA for the TL-CD in Figure~\ref{fig:case-study-2-tlcd-2}}
        \label{fig:case-study-2-causal-dfa-2}
    \end{subfigure}
    \caption{TL-CD for the Cooperative Buttons Task (Left) and respective Causal DFA (Right).}
    \label{fig:case-study-2-tlcd-causal-dfa-2}
\end{figure}
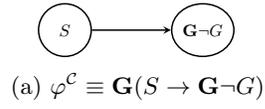
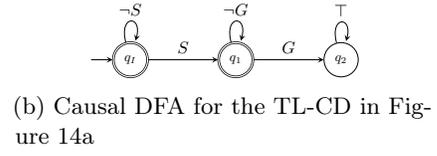

In the cooperative button task, states labeled with S indicate that the agent is in a region from which it cannot proceed to the red button and, consequently, cannot reach the goal.
These labels, called 'Signal lights', serve as indicators of non-goal-achievable states, providing critical information about the agent's position within the environment.
This knowledge is encoded in Figure~\ref{fig:case-study-2-tlcd-causal-dfa-2}, which represents the TL-CD for the cooperative button task. The TL-CD visually outlines the causal pathways and dependencies between states, highlighting the impact of signal lights on the agent's progress. Complementing this diagram, the related Causal DFA formalizes these causal relationships and transitions, aiding in understanding and predicting the agent's behavior within the task.

\end{document}